\documentclass[final]{alt2023} 

\title[Complexity Analysis of a Countable-armed Bandit Problem]{Complexity Analysis of a Countable-armed Bandit Problem}
\usepackage{times}

 \altauthor{\Name{Anand Kalvit} \Email{akalvit22@gsb.columbia.edu}\and
  \Name{Assaf Zeevi} \Email{assaf@gsb.columbia.edu}\\
  \addr Columbia University}

\usepackage{tikz}
\usepackage{enumerate, enumitem, verbatim, bbm, color}
\usepackage{algorithm}
\usepackage[noend]{algpseudocode}
\usepackage{amsmath, amssymb, amsfonts, graphicx}
\usepackage{parskip}
\usepackage{caption} 

\makeatletter
\def\BState{\State\hskip-\ALG@thistlm}
\makeatother

\begin{document}

\maketitle

\begin{abstract}%

We consider a stochastic multi-armed bandit (MAB) problem motivated by ``large'' action spaces, and endowed with a population of arms containing exactly $K$ arm-types, each characterized by a distinct mean reward. The decision maker is oblivious to the statistical properties of reward distributions as well as the population-level distribution of different arm-types, and is precluded also from observing the type of an arm after play. We study the classical problem of minimizing the expected cumulative regret over a horizon of play $n$, and propose algorithms that achieve a rate-optimal finite-time instance-dependent regret of $\mathcal{O}\left( \log n \right)$. We also show that the instance-independent (minimax) regret is $\tilde{\mathcal{O}}\left( \sqrt{n} \right)$ when $K=2$. While the order of regret and complexity of the problem suggests a great degree of similarity to the classical MAB problem, properties of the performance bounds and salient aspects of algorithm design are quite distinct from the latter, as are the key primitives that determine complexity along with the analysis tools needed to study them. 

\end{abstract}


\section{Introduction}

\textbf{Background and motivation.} The MAB model is a classical machine learning paradigm capturing the essence of the \textit{exploration-exploitation} trade-offs characteristic of sequential decision making problems under uncertainty. In its simplest formulation, the decision maker (DM) must play at each time instant $t\in\{1,...,n\}$ one out of a set of $K \ll n$ possible alternatives (\emph{aka} arms), each characterized by a distribution of rewards. Oblivious to their statistical properties, the DM must play a sequence of $n$ arms so as to maximize her cumulative expected payoffs, an objective often converted to minimizing \textit{regret} relative to an oracle with perfect ex ante knowledge of the best arm.

Since the seminal paper of \cite{lai} that laid the main theoretical foundations, there has been a plethora of work developing more advanced MAB models to encapsulate more realistic data-driven decision processes. These include formulations with covariate or contextual information, choice-models, budget constraints, non-stationary rewards, and metric space embeddings, among many others that utilize some structure in the arms, reward distributions, or physics of the problem (see \cite{slivkins2019introduction,lattimore2020bandit} for a comprehensive survey). In this paper, we are motivated by the \emph{choice overload} phenomenon pervading modern MAB applications with a prohibitively large action space such as those encountered in online marketplaces, matching platforms and the likes.

\textbf{Modeling choice overload.} In several applications of MAB, it is quite common for the number of arms to be ``large" to the extent that it may potentially exceed even the horizon of play, i.e. $K \geqslant n$. For example, the problem faced by recommendation systems in large retail platforms, such as Amazon, is characterized by a prohibitively large number of arms (products of certain type) and limited ``display space,'' creating a very challenging combinatorial problem (see, e.g., \cite{mnl-bandit}). Naturally, the canonical MAB model is ill-suited for the study of such settings. Among problems of this nature, a simple yet illuminating abstraction is one where an infinite population of arms is partitioned into $K$ different \emph{arm-types}, each characterized uniquely by some reward statistic (e.g., the mean), and the fraction of each arm-type in the population of arms (or the arm-reservoir) remains fixed over the horizon of play. The motivation to study such settings stems from several contemporary applications, e.g., in a prototypical task-matching problem arising in the online gig economy: the platform must choose upon each task arrival, one agent from a large pool of available agents characterized by unknown (or only partially known) skill proficiencies. Such settings arise naturally in populations endowed with latent \emph{low-dimensional} representations, i.e., an agent can only belong to one of finitely many possible \emph{types}, each characterized distinctly by some attribute. Market segmentation based on types is central also to the operations research literature, see, e.g., \cite{bui2012clustered,banerjee2017segmenting,johari2021matching}, etc., for examples involving analyses of online service and recommendation systems, among several other areas.

\textbf{The countable-armed bandit problem.} We provide an abstraction of the aforementioned decision problem as a bandit with countably many arms, each queried from an infinite population of arms (\emph{aka} the arm-reservoir). There are $K$ possible arm-types given by $\mathcal{K} := \{1,...,K\}$ with $K$ known a priori\footnote{It is routine for platforms to run pilot experiments during initial rounds to gather information on key primitives such as the size and stability of clusters, if any exist in the population. One can therefore safely assume in settings where such clusters strongly exist that the number $K$ of possible types is accurately estimated.}, and the probability vector ${\boldsymbol\alpha} = \left\lbrace \alpha_i : i \in \mathcal{K} \right\rbrace$ denotes their corresponding fraction, i.e., relative prevalence in the reservoir, which is unknown.

Intuitively, the statistical complexity of regret minimization in the simplest formulation of the countable-armed bandit (CAB) with $K=2$ arm-types is governed by three principal primitives: (i) the sub-optimality gap $\underline{\Delta} := \mu_1 - \mu_2$ between the mean rewards $\mu_1 > \mu_2$ of the optimal and inferior arm-types; (ii) the probability $\alpha_1$ of sampling from the population an arm of the optimal type; and (iii) the horizon of play $n$. Absent knowledge of $\left(\underline{\Delta},\alpha_1\right)$, exploration is challenging owing to the ``large'' number of arms. In particular, in contrast with the classical two-armed MAB, in the CAB problem, any finite selection of arms may only contain the mean $\mu_2$. Consequently, any algorithm limited to such a consideration set will suffer a linear regret. Absence of information on $\alpha_1$ further exacerbates the challenges in the study of the CAB problem.


\textbf{Contributions.} There has been limited technical development in this area and the literature remains sparse. In this work, we resolve several foundational questions pertaining to the complexity of this problem setting and provide a comprehensive understanding of various other aspects thereof. Our theoretical contributions can be projected along the following axes:

\textbf{(i) Complexity of regret.} We establish information-theoretic performance lower bounds that are order-wise tight (in the horizon $n$) in the instance-dependent setting (Theorem~\ref{thm:lower_bound_info}). In the instance-independent (minimax) setting, we answer affirmatively an open question on achievability of $\tilde{\mathcal{O}}\left( \sqrt{n} \right)$ regret when $K=2$ and show that this order is best achievable up to poly-logarithmic factors in $n$ (Theorem~\ref{thm:alg1}). In addition, we provide a \emph{uniform} lower bound on achievable performance that is tight in $n$ and explicitly captures the scaling behavior w.r.t. the fraction $\alpha_1$ of optimal arms, and furthermore, has a novel non-information-theoretic proof based entirely on convex analysis (Theorem~\ref{thm:lower_bound0}). 

\textbf{(ii) Algorithm design.} We design algorithms that achieve aforementioned regret guarantees relying only on knowledge of $K$ and are agnostic to information pertaining to the reward distributions as well as the frequency of occurrence of different types. Our design principles (Algorithms~\ref{alg:ALG1} and \ref{alg:ALG2}) are functionally distinct from extant work on finite-armed bandits which reflects in a fundamentally different scaling of regret (see Theorems~\ref{thm:alg1} and \ref{thm:alg2}). We also provide resolution to an outstanding design issue in extant literature for $K=2$ (see Algorithm~\ref{alg:ALG3} and Theorem~\ref{thm:alg3}).

\textbf{(iii) Regret behavior and arm-type distribution.} In contrast to some observations on related models in the literature that show a higher order than $\log n$ (instance-dependent) regret-behavior w.r.t. ${\boldsymbol\alpha}$, we establish that when the learner has knowledge of $K$ but not of ${\boldsymbol\alpha}$, one can still achieve $\mathcal{O}\left( \log n \right)$ regret where the dependence on ${\boldsymbol\alpha}$ only manifests as an additive loss (Theorems~\ref{thm:alg2} and \ref{thm:alg3}).

Before proceeding with a formal description of our model, we provide a brief overview of related works below.

\textbf{Extant literature on bandits with infinitely many arms.} These problems involve an infinite population of arms and a fixed \emph{reservoir distribution} over a (typically uncountable) set of arm-types; a common reward statistic (usually the mean) uniquely characterizes each arm-type. The infinite-armed bandit problem traces its roots to \cite{berry1997bandit} where the problem was studied under Bernoulli rewards and a reservoir distribution of Bernoulli parameters that is Uniform on $[0,1]$. Subsequent works have considered more general reward and reservoir distributions on $[0,1]$, see, e.g., \cite{wang2009,bonald2013two,carpentier2015simple,chan2018infinite}. In terms of the statistical complexity of regret minimization, an uncountably rich set of arm-types is tantamount to the minimal achievable regret being polynomial in the horizon of play (see aforementioned references). In contrast, the recently studied models in \cite{kalvit2020finite,de2021bandits} that our work is most closely related to, are fundamentally simpler owing to a finite set of arm-types; this is central to the achievability of logarithmic (instance-dependent) regret in this class of problems. These two works are briefly discussed below. 

The CAB problem first appeared in \cite{kalvit2020finite} together with an online adaptive policy achieving $\mathcal{O}\left( \log n \right)$ regret when $K=2$. This policy is derived from UCB1 \citep{auer2002} and relies on certain newly discovered concentration and convergence properties thereof (see \cite{kalvit2021closer} for a detailed discussion of said properties). However, the analysis of this policy cannot be adapted to $K>2$ types; we will later provide an example with $K>2$ where the policy will likely run into issues that can be effectively mitigated by the algorithms proposed in this paper. There is also recent literature \citep{de2021bandits} on a related setting where the set of inferior arm-types may be arbitrary as long as it is $\underline{\Delta}$-separated from the optimal mean. However, ex ante knowledge of the proportion $\alpha_1$ of optimal arms is necessary to achieve logarithmic regret in this setting. This aspect distinguishes their setting from CAB and will be discussed at length later. 

Lastly, a formulation of the countable-armed problem based on \emph{pure exploration}, referred to as the ``heaviest coin identification problem,'' was studied in \cite{chandrasekaran2014finding} for $K=2$ (see \cite{jamieson2016detection} for subsequent developments). In contrast, our problem is based on optimization of cumulative payoff (or regret); as a result, it shares little similarity with cited works.



\textbf{Outline of the paper.} A formal description of the model is provided in \S\ref{sec:formulation}; \S\ref{sec:lower_bound} discusses lower bounds on achievable performance. We propose our algorithms in \S\ref{sec:algorithms} and state supporting theoretical guarantees. Numerical experiments, proofs, and auxiliary results are relegated to the appendix.

\section{Problem formulation}
\label{sec:formulation}

The set of arm-types is given by $\mathcal{K}=\left\lbrace 1,...,K \right\rbrace$, and the decision maker (DM) only knows the cardinality $K$ of $\mathcal{K}$. Each type $i\in\mathcal{K}$ is characterized by a unique mean reward $\mu_i$; the reservoir is thus characterized by the collection ${\boldsymbol\mu} := \left\lbrace \mu_i : i\in\mathcal{K} \right\rbrace$ of possible mean rewards. Without loss of generality, we assume $\mu_1 > ... > \mu_K$ and refer to type~$1$ as the \emph{optimal} type (we may refer to the others as inferior types). Define $\bar{\Delta} := \mu_1 - \mu_K$ and $\underline{\Delta} := \mu_1 - \mu_2$ as the maximal and minimal sub-optimality gaps respectively, and $\delta := \min_{1 \leqslant i < j \leqslant K}\left( \mu_i - \mu_j \right)$ as the minimal reward gap. Finally, ${\boldsymbol\alpha} := \left( \alpha_i : i\in\mathcal{K} \right)$ denotes the vector of reservoir probabilities for each type (\emph{aka} the reservoir distribution), coordinate-wise bounded away from $0$. These primitives will be important drivers of the statistical complexity of the regret minimization problem, as we shall later see. The horizon of play is $n$, and the DM must play one arm at each time $t\in\{1,...,n\}$. 

The set of arms that have been played up to and including time $t\in\{1,2,...\}$ is denoted by $\mathcal{I}_t$ (and $\mathcal{I}_0 := \phi$). The set of actions available to the DM at time $t$ is given by $\mathcal{A}_t= \mathcal{I}_{t-1}\cup\{\text{new}_t\}$; the DM must either play an arm from $\mathcal{I}_{t-1}$ at time $t$ or select the action ``$\text{new}_t$'' which corresponds to querying (and playing) a new arm from the reservoir. This new arm is optimal-typed with probability $\alpha_1$ and sub-optimal otherwise. The DM is oblivious to ${\boldsymbol\alpha}$, and furthermore precluded from observing the type of an arm upon query or play. A policy $\pi:=\left(\pi_1, \pi_2, ...\right)$ is an adaptive allocation rule that prescribes at time $t$ an action $\pi_t$ from $\mathcal{A}_t$ (possibly randomized). Each pull (or play) of an arm results in a stochastic reward. The sequence of rewards realized from the first $k$ pulls of an arm labeled $i$ (henceforth called arm~$i$) is denoted by $\left( X_{i,j} \right)_{1\leqslant j \leqslant k}$; these are mean-preserving in time keeping the arm fixed, independent across arms and time, and take values in $[0,1]$. The natural filtration at time $t$, denoted by $\mathcal{F}_t$ and defined w.r.t. the sequence of rewards realized up to and including time $t$, is given by $\mathcal{F}_t := \sigma\left\lbrace \left( \pi_s \right)_{1\leqslant s \leqslant t}, \left\lbrace \left( X_{i,j} \right)_{ 1 \leqslant j \leqslant N_i(t)} : i\in\mathcal{I}_t \right\rbrace \right\rbrace$ (with $\mathcal{F}_0:=\phi$), where $N_i(t)$ denotes the number of pulls of arm~$i$ up to and including time $t$. The cumulative \emph{pseudo-regret} of policy $\pi$ after $n$ plays is given by $R_n^{\pi}:=\sum_{t=1}^{n}\left(\mu_1 - \mu_{\mathcal{T}\left(\pi_t\right)}\right)$, where $\mathcal{T}\left(\pi_t\right)\in\mathcal{K}$ denotes the type of the arm played by $\pi$ at time $t$; note that $R_n^{\pi}$ is a sample path-dependent quantity. The DM is interested in the classical problem of minimizing the expectation of the \emph{cumulative regret} $\hat{R}_n^\pi := \sum_{t=1}^{n}\left( \mu_1 - X_{\pi_t,N_{\pi_t}(t)} \right)$, given by
\begin{align}
\inf_{\pi\in\Pi}\mathbb{E}\hat{R}_n^\pi = \inf_{\pi\in\Pi}\mathbb{E}\left[ \sum_{t=1}^{n}\left( \mu_1 - X_{\pi_t,N_{\pi_t}(t)} \right) \right] \underset{\mathrm{(\dag)}}{=} \inf_{\pi\in\Pi}\mathbb{E}\left[ \sum_{t=1}^{n}\left(\mu_1 - \mu_{\mathcal{T}\left(\pi_t\right)}\right) \right] = \inf_{\pi\in\Pi}\mathbb{E}R_n^\pi, \label{eqn:problem_formulation}
\end{align}
where $(\dag)$ follows from the Tower property of expectation, the infimum is over policies satisfying the \emph{non-anticipation} property $\pi_t : \mathcal{F}_{t-1}\to \mathcal{P}\left( \mathcal{A}_{t} \right)$ for $t\in\{1,2,...\}$; $\mathcal{P}\left( \mathcal{A}_{t} \right)$ denotes the probability simplex on $\mathcal{A}_t$. Accordingly, the expectation in \eqref{eqn:problem_formulation} is w.r.t. all the possible sources of randomness in the problem (rewards, policy, and the arm-reservoir). 

\section{Lower bounds for natural policy classes}
\label{sec:lower_bound}

There are three fundamental primitives governing the complexity of achievable regret in this setting, viz., (i) the minimal sub-optimality gap $\underline{\Delta}$; (ii) the proportion $\alpha_1$ of the optimal arm-type in the reservoir; and (iii) the number of arm-types $K$. We next characterize lower bounds on achievable performance w.r.t. each of these primitives.

\subsection{Achievable performance w.r.t. $\boldsymbol{\underline{\Delta}}$: Information-theoretic lower bounds}

The statistical complexity of this problem setting is best illustrated via the paradigmatic case of $K=2$ and $\alpha_1 \leqslant 1/2$. In this case, one anticipates the problem to be at least as hard as the classical two-armed bandit with a mean reward gap of $\underline{\Delta}$. Indeed, we establish this in Theorem~\ref{thm:lower_bound_info} via information-theoretic reductions adapted to handle a countable number of arms (proof is provided in Appendix~\ref{appendix:lower_bound_info}). In what follows, an instance $\nu$ of the problem refers to a collection of reward distributions with gap $\underline{\Delta}$; note that we are excluding the reservoir probabilities $\boldsymbol\alpha = \left( \alpha_1, \alpha_2 \right)$ from the definition of an instance. Recall that $\alpha_1$ denotes the reservoir probability associated with the optimal mean reward, and $\alpha_2 = 1 - \alpha_1$ is the probability of the inferior. We will overload the notation for expected cumulative regret slightly to emphasize its dependence on $\nu$ as well as $\boldsymbol{\alpha}$.

\begin{definition}[Admissible policies when $\boldsymbol{ K=2}$]
\label{def:consistency}
A policy $\pi$ is deemed admissible if for any instance $\nu$, reservoir distributions $\boldsymbol\alpha=\left( \alpha_1,1-\alpha_1 \right), \boldsymbol\alpha^\prime=\left( \alpha_1^\prime,1-\alpha_1	^\prime \right)$, and horizon $n$, one has that $\mathbb{E}R_n^\pi\left( \nu, \boldsymbol\alpha^\prime \right) \geqslant \mathbb{E}R_n^\pi\left( \nu, \boldsymbol\alpha \right)$ whenever $\alpha_1^\prime \leqslant \alpha_1$. The set of such policies is denoted by $\Pi_{\texttt{adm}}$.
\end{definition}

We remark that the aforementioned definition is not restrictive in our problem setting since it is only natural that any reasonable policy should incur a larger cumulative regret (in expectation) in problems where the reservoir holds fewer optimal arms (in proportion).

\begin{theorem}[Information-theoretic lower bounds when $\boldsymbol{ K=2}$]
\label{thm:lower_bound_info}
There exists an absolute constant $C>0$ such that the following holds under any $\pi\in\Pi_{\texttt{adm}}$ and any $\boldsymbol\alpha$ with $\alpha_1 \leqslant 1/2$:
\begin{enumerate}
\item For any $\underline{\Delta}>0$, there exists a problem instance $\nu$ such that $\mathbb{E}R_n^\pi\left( \nu, \boldsymbol\alpha \right) \geqslant C\log n/\underline{\Delta}$ for large enough $n$, where the ``large enough $n$'' depends exclusively on $\underline{\Delta}$.
\item For any $n\in\mathbb{N}$, there exists a problem instance $\nu$ such that $\mathbb{E}R_n^\pi\left( \nu, \boldsymbol\alpha \right) \geqslant C\sqrt{n}$.
\end{enumerate}
\end{theorem}

\textbf{Distinction from classical MAB.} Although the above result bears resemblance to classical information-theoretic lower bounds for finite-armed bandits, it is imperative to note that the setting has a fundamentally greater complexity that requires a more nuanced analysis vis-\`a-vis the finite-armed problem. Traditional proofs, as a result, cannot be tailored to our setting in a translational manner. To see this, note that when $\alpha_1$ is high, a query of the reservoir is very likely to return an arm of the optimal type; in the limit as $\alpha_1\to 1$, the problem becomes degenerate as all policies incur zero expected regret. Clearly, the problem cannot be harder than a two-armed bandit with gap $\underline{\Delta}$ uniformly over all values of $\alpha_1$. While we conjecture $\alpha_1<1$ to be a sufficient condition for the existence of $\Omega\left( \log n/\underline{\Delta} \right)$ instance-dependent and $\Omega\left( \sqrt{n} \right)$ instance-independent (minimax) lower bounds, there are technical challenges due to probabilistic type associations over countably many arms. The restriction to $\alpha_1 \leqslant 1/2$ and \emph{admissible} policies (Definition~\ref{def:consistency}) is then necessary for tractability of the proof and it remains unclear if this can be generalized further. Furthermore, when $K>2$, even defining an appropriate notion of admissibility \`a la Definition~\ref{def:consistency} is non-trivial and will likely involve dependencies on $\boldsymbol\mu$ in addition to $\boldsymbol\alpha$; pursuits in this direction are currently left to future work. 

\subsection{Achievable performance w.r.t. $\boldsymbol{\alpha_1}$: A uniform lower bound for front-loaded policies} 

Though the bounds in Theorem~\ref{thm:lower_bound_info} are tight in $n$ as we shall later see, they fail to provide any actionable insights w.r.t. ${\boldsymbol\alpha}$. A natural question in the CAB setting is whether and in what manner does the presence of countably many arms affect achievable regret. In particular, how does the difficulty associated with finding optimal arms from the reservoir (and the dependence on the distribution ${\boldsymbol\alpha}$) come into play. Below, we propose a lower bound that explicitly captures this dependence, albeit with respect to a somewhat restricted policy class. 

\begin{theorem}[$\boldsymbol{\alpha}$-dependent lower bound for any $\boldsymbol{K\geqslant 2}$]
\label{thm:lower_bound0}
Denote by $\tilde{\Pi}$ the class of policies under which the decision to query the arm-reservoir at any time $s\in\{1,2,...\}$ is independent of $\mathcal{F}_{s-1}$. Then, for all problem instances $\nu$ with a minimal sub-optimality gap of at least $\underline{\Delta}>0$, one has
\begin{align*}
\liminf_{n\to\infty}\inf_{\pi\in\tilde{\Pi}}\frac{\mathbb{E}R_n^\pi\left( \nu, {\boldsymbol\alpha} \right)}{\log n} \geqslant \frac{\left(1-\alpha_1\right)^2\underline{\Delta}}{\alpha_1}.
\end{align*}
\end{theorem}

\textbf{Discussion.} The proof is located in Appendix~\ref{appendix:lower_bound0}. Several comments are in order. \textbf{(i)} The class $\tilde{\Pi}$, in particular, includes policies that front-load queries, i.e., query the reservoir upfront for a pre-specified number of arms and then deploy a regret minimizing routine on the queried set until the end of the playing horizon, see, e.g., the \texttt{Sampling-UCB} algorithm due to \cite{de2021bandits}. \textbf{(ii)} The cited paper also derives an information-theoretic $\Omega\left( \log n/\alpha_1\underline{\Delta} \right)$ lower bound based on a standard reduction to a hypothesis testing problem, although notably their setting is non-trivially distinct from ours (this reflects starkly different sensitivities of achievable regret to $1/\alpha_1$-scale, as we shall later see). Importantly though, akin to Theorem~\ref{thm:lower_bound_info}, their bound too, establishes the \emph{existence} of an instance with logarithmic regret. On the other hand, the foremost noticeable aspect of Theorem~\ref{thm:lower_bound0} that differs from aforementioned results is that it provides a uniform lower bound \emph{over all instances} that are at least $\underline{\Delta}$-separated in the mean reward, as opposed to merely establishing their existence. \textbf{(iii)} The presence of $\underline{\Delta}$ in the numerator (unlike traditional bounds where $\underline{\Delta}$ resides in the denominator) suggests that while this bound may be vacuous if $\underline{\Delta}$ is ``small,'' it certainly becomes most relevant when $\underline{\Delta}$ is ``well-separated.'' At the same time, it should be noted that Theorem~\ref{thm:lower_bound0} does not contradict the $\mathcal{O}\left( \log n/\alpha_1\underline{\Delta} \right)$ upper bound (up to logarithmic factors in $\underline{\Delta}$) of \texttt{Sampling-UCB}; it merely provides a tool to separate regimes of $\underline{\Delta}$ where one bound captures the dominant effects vis-\`a-vis the other. \textbf{(iv)} A novelty of Theorem~\ref{thm:lower_bound0} lies in its proof, which differs from classical lower bound proofs in that it is based entirely on ideas from convex analysis as opposed to the information-theoretic and change-of-measure techniques hitherto used in the literature.

\textbf{Remarks}. \textbf{(i)} It is not impossible to avoid the $1/\alpha_1$-scaling of the instance-dependent logarithmic regret. We will later show via an upper bound for one of our algorithms (\texttt{ALG2}$(n)$) that the $\alpha_1$-dependence can, in fact, be relegated to constant order terms (\texttt{ALG2}$(n)$ queries arms \emph{adaptively} based on sample-history and therefore does not belong to $\tilde{\Pi}$). Importantly, this will somewhat surprisingly establish that the instance-dependent logarithmic bound in Theorem~\ref{thm:lower_bound_info} is optimal w.r.t. to its dependence on $\alpha_1$ (the scaling w.r.t. $\underline{\Delta}$, however, may not be best possible as forthcoming upper bounds suggest). \textbf{(ii)} Theorem~\ref{thm:lower_bound0} holds also for any arm-reservoir where the optimal type is at least $\underline{\Delta}$-separated from the rest, the nature of types (countable or uncountable) notwithstanding.

\subsection{Achievable performance w.r.t. $\boldsymbol K$: The Bandit and the Coupon-collector}

In the classical $K$-armed bandit problem, the (instance-dependent) regret scales linearly with the number of arms. We will next show that the $K$-typed countable-armed setting studied in this paper differs from its $K$-armed counterpart on account of a fundamentally distinct scaling of regret w.r.t. $K$. We will illustrate this by pivoting to a full information setting with one-sample learning, i.e., a setting where the decision maker observes the mean reward of an arm immediately upon pulling it, but does not learn whether it is optimal. Under such a setting, the optimal policy $\pi^\ast$ for the $K$-armed problem will pull each of the $K$ arms once and subsequently commit the residual budget of play to the optimal arm, thus incurring a \emph{lifetime regret} of $\mathbb{E}R_\infty^{\pi^\ast} = \sum_{i=2}^{K}\left( \mu_1 - \mu_i \right) = \Theta(K)$. The optimal policy for the $K$-typed countable-armed setting will, analogously, keep querying new arms from the reservoir until it has collected one of each of the $K$ types, and will subsequently commit to the arm within said collection that has the best mean reward. In this case, regret will only accrue until the decision maker has pulled one arm of each type.

\begin{theorem}[Regret scaling w.r.t. $\boldsymbol K$]
\label{thm:regret-K}
In the full information setting, the lifetime regret of any policy under reservoir distribution $\boldsymbol\alpha$ and mean reward vector $\boldsymbol\mu$ is at least $\sum_{i=2}^{K}\alpha_i\left( \mu_1 - \mu_i \right)K\log K$.
\end{theorem}

If the reservoir distribution remains non-degenerate w.r.t. the optimal type, i.e., the fraction $\alpha_1$ of optimal arms in the reservoir remains bounded away from $1$ as $K$ increases, it is ensured that $\sum_{i=2}^{K}\alpha_i\left( \mu_1 - \mu_i \right)$ remains non-vanishing in $K$. Consequently, the lower bound in Theorem~\ref{thm:regret-K} grows as $\Omega\left( K\underline{\Delta}\log K \right)$.

This result establishes a fundamentally distinct scaling of regret w.r.t. $K$ in the countable-armed setting vis-\`a-vis the $K$-armed one (in the full information setting). When the true type of an arm is not immediately observable, one only expects the $\Omega\left( K\log K \right)$ scaling to exacerbate. In fact, when the learning horizon is $n$, we conjecture that regret grows at least as $\Omega\left( K\log K\log n \right)$, where the $\Omega(\cdot)$ is modulo gap-dependent constants. Characterizing the information-theoretic optimal rate, however, remains a challenging open problem. The proof of Theorem~\ref{thm:regret-K} is provided in Appendix~\ref{proof:regret-K}.

\section{Gap and reservoir adaptive policies}
\label{sec:algorithms}

As discussed, our goal here is to investigate regret achievable under ${\boldsymbol\mu}$-adaptive algorithms that are agnostic also to ex ante information on the distribution ${\boldsymbol\alpha}$ of possible arm types. We propose two algorithms; \texttt{ALG1}$(n)$ and \texttt{ALG2}$(n)$, that are both predicated on ex ante knowledge of the horizon of play $n$. \S\ref{subsec:fixed} discusses the first of these, \texttt{ALG1}$(n)$, which uses knowledge of $n$ to calibrate the duration of its exploration phases. \texttt{ALG1}$(n)$ serves as an insightful basal motif for algorithm design in that it satisfies the desiderata of an $\mathcal{O}\left( \log n \right)$ instance-dependent regret for general $K \geqslant 2$ as well as an $\tilde{\mathcal{O}}\left( \sqrt{n} \right)$ instance-independent (minimax) regret when $K=2$; the latter property settling an open problem in the literature. However, its regret has a sub-optimal dependence on ${\boldsymbol\alpha}$. We leverage structural insights from the analysis of \texttt{ALG1}$(n)$ to explore another design in \texttt{ALG2}$(n)$ in \S\ref{subsec:adaptive}, which determines its exploration phase lengths \emph{adaptively}, as opposed to pre-specifying them upfront. This new design guarantees the best possible dependence of regret on ${\boldsymbol\alpha}$. Finally in \S\ref{subsec:UCB}, we discuss a fully sequential adaptive algorithm from extant literature for $K=2$, and propose a simple modification to rid it of a certain fragile assumption pertaining to ex ante knowledge of the support of reward distributions. We also provide new sharper bounds for the modified algorithm and discuss potential issues with its generalization to $K \geqslant 2$ vis-\`a-vis \texttt{ALG1}$(n)$ and \texttt{ALG2}$(n)$. 

\subsection{Explore-then-commit with a pre-specified exploration schedule}
\label{subsec:fixed}

In what follows (and all subsequent algorithms), a \emph{new} arm is one that is freshly queried from the reservoir (an arm without a history of previous pulls). This arm belongs to type~$i$ with probability $\alpha_i$ independent of the problem history thus far (collection of arms and types queried and the corresponding reward realizations until the current time).

\begin{algorithm}
\caption{\texttt{ALG1}$\left(n\right)$ \texttt{(Fixed design ETC)}}
\label{alg:ALG1}
\begin{algorithmic}[1]
\BState \textbf{Input:} Horizon of play $n$.
\State Set budget $T=n$; set epoch counter $k=1$.
\BState \textbf{Initialize new epoch:} Query $K$ \emph{new} arms; call it consideration set $\mathcal{A}=\{1,...,K\}$.
\State Set exploration duration $L = \left\lceil e^{2\sqrt{k}} \log n\right\rceil$.
\State $m \gets \min \left( L, \left\lfloor T/K \right\rfloor \right)$.
\State Play each arm in $\mathcal{A}$ $m$ times; observe rewards $\left\lbrace \left( X_{1,j}, ..., X_{K,j}\right) : j=1,...,m \right\rbrace$.
\State Update budget: $T \gets T - Km$.
\If {$\exists$ $a,b\in\mathcal{A}$, $a<b$ s.t. $\left\lvert {\sum_{j=1}^{m}(X_{a,j}-X_{b,j})} \right\rvert < 2me^{-\sqrt{k}}$}
\State Permanently discard $\mathcal{A}$.
\State $k \gets k + 1$.
\State Repeat from step (3).
\Else
\State Permanently commit to arm $a^{\ast} \in \arg\max_{a\in\mathcal{A}}\left\lbrace \sum_{j=1}^{m}X_{a,j} \right\rbrace$.
\EndIf
\end{algorithmic}
\end{algorithm}

\textbf{Discussion of Algorithm~\ref{alg:ALG1}.} The foremost noticeable feature of this algorithm is the (nearly) exponentially increasing exploration schedule. Specifically, in the $k$\textsuperscript{th} epoch, each of the $K$ arms in the consideration set is played $\left\lceil e^{2\sqrt{k}}\log n \right\rceil$ times. It suffices to cap the size of the consideration set at $K$ since the decision maker is a priori aware of the existence of exactly $K$ arm-types in the reservoir. Upon completion of the $k$\textsuperscript{th} epoch, the cumulative-difference-of-reward statistic for each of the ${K \choose 2}$ arm-pairs is compared against a threshold of $2e^{-\sqrt{k}}m$, where $2e^{-\sqrt{k}}$ should be imagined as a proxy for a lower bound on the minimal reward gap $\delta$. If said statistic is small relative to the threshold for some pair, the pair is likely to contain arms of the same type (equal means), in which case, the algorithm discards the \emph{entire} consideration set and ushers in a new epoch with a larger exploration phase. This is done to avoid the possibility of incurring linear regret should an optimal arm be missing from the consideration set (e.g., when all $K$ arms belong to type~$2$). On the other hand, if all arm-pairs are sufficiently separated, the algorithm simply commits permanently to the empirically best arm. The intuition behind the (nearly) exponential schedule is that as $k$ grows, $2e^{-\sqrt{k}}$ will eventually provide a lower bound on $\delta$, and one may hope to achieve appropriate levels of error control using window sizes in the $k$\textsuperscript{th} epoch. Full proof is provided in Appendix~\ref{proof:alg1}.

\begin{theorem}[Upper bound on the regret of \texttt{\textbf{ALG1}}$\boldsymbol{(n)}$]
\label{thm:alg1}
For a horizon of play $n \geqslant K$, the expected cumulative regret of the policy $\pi$ given by \texttt{ALG1}$(n)$ is bounded as
\begin{align*}
\mathbb{E}R_n^{\pi} \leqslant \frac{\tilde{C}_{{\boldsymbol\alpha}}\bar{\Delta}\log n}{\delta^2}\log^2\left( \frac{4}{\delta} \right) + 2K\bar{\Delta},
\end{align*}
where $\tilde{C}_{{\boldsymbol\alpha}}$ is some constant that depends only on ${\boldsymbol\alpha}$; an exact expression is provided in \eqref{eqn:C_alpha_K}. In particular, $\tilde{C}_{{\boldsymbol\alpha}}\to\infty$ as $\prod_{i=1}^{K}\alpha_i$ approaches $0$.
\end{theorem}

\textbf{Discussion.} The dependence on the \emph{minimal reward gap} $\delta$ in Theorem~\ref{thm:alg1} is not an artifact of our analysis but, in fact, reflective of the operating principle of the algorithm. \texttt{ALG1}$(n)$ keeps querying new consideration sets of size $K$ until it determines with high enough confidence that the queried arms are all distinct-typed; this is the genesis of $\delta$ in the upper bound. Importantly, equipped just with knowledge of $K$, it remains unclear if there exists an alternative sampling strategy that does not rely on assessing pairwise similarities between the queried arms, without necessitating any additional information on ${\boldsymbol\alpha}$. Furthermore, while \texttt{ALG1}$(n)$ is evidently rate-optimal in $n$ (in the instance-dependent sense), the scaling of its upper bound w.r.t. ${\boldsymbol\alpha}$ is far from optimal. In particular, the ${\boldsymbol\alpha}$-dependent factor in the leading term is attributable to a naive pre-determined exploration schedule. This dependence can, in fact, be relegated to $\mathcal{O}(1)$ terms using a more sophisticated policy that operates based on an adaptive determination of stopping and re-initialization times. 

\textbf{Remarks.} \textbf{(i)} When $K=2$, the upper bound in Theorem~\ref{thm:alg1} reduces to $\left(\tilde{C}_{{\boldsymbol\alpha}}/\underline{\Delta}\right)\log^2\left( {4}/{\underline{\Delta}} \right)\log n + 4\underline{\Delta}$, leading to a worst-case regret (w.r.t. $\underline{\Delta}$) of $\tilde{\mathcal{O}}\left( \tilde{C}_{{\boldsymbol\alpha}} \sqrt{n} \right)$, where the big-Oh only hides poly-logarithmic factors in $n$. This settles an open question concerning the best achievable minimax regret in the countable-armed problem with two types.\footnote{Minimax guarantees in previous work were polynomially bounded away from $\sqrt{n}$; see \cite{kalvit2020finite}.} \textbf{(ii)} While specifying the exploration schedule, the choice of the exponent in $k$ can be fairly general as long as it is coercive and grows sufficiently fast but sub-linearly; the square-root function is chosen for technical convenience. Instead, if one were to use a linear function of $k$ in the exponent, the algorithm's performance would become fragile w.r.t. ex ante knowledge of ${\boldsymbol\alpha}$; an ill-calibrated \texttt{ALG1}$(n)$ can potentially incur linear regret.

\subsection{Explore-then-commit with an adaptive stopping time}
\label{subsec:adaptive}

\begin{algorithm}
\caption{\texttt{ALG2}$\left(n\right)$ \texttt{(ETC with adaptive stopping times)}}
\label{alg:ALG2}
\begin{algorithmic}[1]
\BState \textbf{Input:} Horizon of play $n$.
\BState Set budget $T=n$; Burn-in samples (per arm) $s_n$.
\BState \textbf{Initialize new epoch:} Query $K$ \emph{new} arms; call it consideration set $\mathcal{A}=\{1,...,K\}$.
\State Play each arm in $\mathcal{A}$ for $s_n$ periods; observe rewards $\left\lbrace X_{a,j} : a\in\mathcal{A}, j=1,...,s_n\right\rbrace$. 
\State Set per-arm sample count $m=s_n$.
\State Update budget: $T \gets T - s_nK$.
\State Generate ${K \choose 2}$ independent standard Gaussian random variables $\left\lbrace \mathcal{Z}_{a,b} : a,b\in\mathcal{A}, a<b \right\rbrace$.
\While {$T \geqslant K$}
	\If {$\exists$ $a,b\in\mathcal{A}$, $a<b$ s.t. $\left\lvert \mathcal{Z}_{a,b} + \sum_{j=1}^{m}\left( X_{a,j}-X_{b,j} \right) \right\rvert < 4\sqrt{m\log m}$}
		\State Permanently discard $\mathcal{A}$ and repeat from step (3).
	\Else
		\If {$\left\lvert \sum_{j=1}^{m}\left( X_{a,j}-X_{b,j} \right) \right\rvert \geqslant 4\sqrt{m\log n}\ \forall\ a,b\in\mathcal{A},\ a<b$}
			\State Permanently commit to arm~$a^\ast\in\arg\max_{a\in\mathcal{A}}\left\lbrace \sum_{j=1}^{m} X_{a,j} \right\rbrace$.
		\Else
			\State Play each arm in $\mathcal{A}$ once; observe rewards $\left\lbrace X_{a,m+1} : a\in\mathcal{A} \right\rbrace$.
			\State $m \gets m+1$.
			\State $T \gets T - K$.
		\EndIf
	\EndIf
\EndWhile
\end{algorithmic}
\end{algorithm}

\textbf{Discussion of Algorithm~\ref{alg:ALG2}.} At any time, the algorithm computes two thresholds; $\mathcal{O}\left( \sqrt{m\log m} \right)$ and $\mathcal{O}\left( \sqrt{m\log n} \right)$ for the ${K \choose 2}$ pairwise difference-of-reward statistics, $m$ being the per-arm sample count. If said statistic is dominated by the former threshold for some arm-pair, it is likely to contain arms of the same type (equal means). The explanation stems from the Law of the Iterated Logarithm (see \cite{durrett2019probability}, Theorem~8.5.2): a zero-drift length-$m$ random walk has its envelope bounded by $\mathcal{O}\left( \sqrt{m\log\log m} \right)$. In the aforementioned scenario, the algorithm discards the \emph{entire} consideration set and ushers in a new one. This is done to avoid the possibility of incurring linear regret should an optimal arm be missing from the consideration set. In the other scenario that the difference-of-reward statistic dominates the larger threshold for all arm-pairs, the consideration set is likely to contain arms of distinct types (no two have equal means) and the algorithm simply commits to the empirically best arm. Lastly, if difference-of-reward lies between the two thresholds (signifying insufficient learning), the sample count for each arm is advanced by one, and the entire process repeats.

\textbf{Reason for introducing zero-mean corruptions supported on $\boldsymbol{\mathbb{R}}$.} Centered Gaussian noise is added to the difference-of-reward statistic in step (9) of \texttt{ALG2}$(n)$ to avoid the possibility of incurring linear regret should the support of the reward distributions be a ``very small'' subset of $[0,1]$. To illustrate this point, suppose that $K=2$, $s_n=1$, and the rewards associated with the two types are deterministic with $\underline{\Delta} < 2\sqrt{2\log 2}$. Then, as soon as the algorithm queries a heterogeneous consideration set (one arm optimal and the other inferior) and the per-arm sample count reaches $2$, the difference-of-reward statistic will satisfy $\left\lvert \sum_{j=1}^{2} \left( X_{1,j} - X_{2,j} \right) \right\rvert = 2\underline{\Delta} < 4\sqrt{2\log 2}$, resulting in the consideration set getting discarded. On the other hand, if the consideration set is homogeneous (both arms simultaneously optimal or inferior), the algorithm will still re-initialize as soon as the per-arm count reaches $2$.\footnote{$\left\lvert \sum_{j=1}^{m} \left( X_{1,j} - X_{2,j} \right) \right\rvert = 0$ identically in this case for any $m\in\mathbb{N}$ while $4\sqrt{m\log m}>0$ only for $m\geqslant 2$.} This will force the algorithm to keep querying new arms from the reservoir at rate that is linear in time, which is tantamount to incurring linear regret in the horizon. The addition of centered Gaussian noise hedges against this risk by guaranteeing that the difference-of-reward process essentially has an infinite support at all times even when the reward distributions might be degenerate. This rids the regret performance of its fragility w.r.t. the support of reward distributions. The next proposition crystallizes this discussion; proof is provided in Appendix~\ref{appendix:propK}.

\begin{proposition}[Persistence of heterogeneous consideration sets]
\label{propK}
Suppose $\left\lbrace X_{a,j} : j=1,2,... \right\rbrace$ is a collection of independent samples from an arm of type~$a\in\{1,...,K\} =: \mathcal{A}$. Let $\left\lbrace \mathcal{Z}_{a,b} : a,b\in\mathcal{A}, a<b \right\rbrace$ be a collection of ${K \choose 2}$ independently generated standard Normal random variables. Then,
\begin{align}
\mathbb{P}\left( \bigcap_{m \geqslant 1} \bigcap_{a,b\in\mathcal{A}, a<b}\left\lbrace \left\lvert \mathcal{Z}_{a,b} + \sum_{j=1}^{m} \left( X_{a,j} - X_{b,j} \right) \right\rvert \geqslant 4\sqrt{m\log m} \right\rbrace \right) > \frac{\bar{\Phi}\left( f\left( T_0 \right) \right)}{2} =: \beta_{\delta,K} > 0, \label{eqn:beta_K} 
\end{align}
where $\bar{\Phi}(\cdot)$ is the right tail of the standard Normal CDF, and $T_0 := \max \left( \left\lceil \left( 64/\delta^2 \right) \log^2\left( \frac{64}{\delta^2} \right) \right\rceil, \Lambda_K \right)$ with $\Lambda_K := \inf \left\lbrace p \in\mathbb{N} : \sum_{m=p}^{\infty}\frac{1}{m^8} \leqslant \frac{1}{2K^2} \right\rbrace$. Lastly, $f(x) := x + 4\sqrt{x\log x}$ for all $x \geqslant 1$.
\end{proposition}

\textbf{Interpretation of $\boldsymbol{\beta_{\delta,K}}$.} First of all, note that $\beta_{\delta,K}$ admits a closed-form characterization in terms of standard functions and satisfies $\beta_{\delta,K} > 0$ for $\delta > 0$ with $\lim_{\delta\to 0}\beta_{\delta,K} = 0$. Secondly, \emph{$\beta_{\delta,K}$ depends exclusively on $\delta$ and $K$}, and represents a lower bound on the probability that \texttt{ALG2}$(n)$ will never discard a consideration set containing arms of distinct types. This meta-result will be key to the upper bound on the regret of \texttt{ALG2}$(n)$ stated next in Theorem~\ref{thm:alg2}. 

\begin{theorem}[Upper bound on the regret of \texttt{\textbf{ALG2}}$\boldsymbol{(n)}$]
\label{thm:alg2}
For a horizon of play $n \geqslant K$ and per-arm burn-in phase of $1 \leqslant s_n \leqslant n/K$ samples, the expected cumulative regret of the policy $\pi$ given by \texttt{ALG2}$(n)$ is bounded as
\begin{align*}
\mathbb{E}R_n^\pi \leqslant \frac{CK^3\bar{\Delta}}{\gamma\left(s_n\right)}\left( \frac{\log n}{{\delta}^2} + \frac{s_n}{K!\prod_{i=1}^{K}\alpha_i} \right),
\end{align*} 
where $C$ is some absolute constant, and $\gamma\left(s_n\right)$ is as defined in \eqref{eqn:gamma-scaling}. In particular, $\gamma(t)$ is monotone increasing in $t$ with $\gamma(t)\to 1$ as $t\to\infty$, and $\gamma\left(1\right) = \beta_{\delta,K}$, where the latter is as defined in \eqref{eqn:beta_K}.
\end{theorem}

\textbf{Remarks.} The dependence on $\delta$ in Theorem~\ref{thm:alg2} is not incidental and has the same genesis as discussed in the context of Theorem~\ref{thm:alg1}. However, there is a prominent distinction from Theorem~\ref{thm:alg1} in that the dependence on ${\boldsymbol\alpha}$ is captured exclusively through the constant term (as opposed to the logarithmic term). This should be viewed in light of the lower bound in Theorem~\ref{thm:lower_bound0}; by allowing for policies that query the arm-reservoir \emph{adaptively}, one can potentially make the regret performance robust w.r.t. ${\boldsymbol\alpha}$. Absence of ${\boldsymbol\alpha}$ from the leading term also leads to the somewhat remarkable conclusion that the lower bound in Theorem~\ref{thm:lower_bound_info} is optimal w.r.t. dependence on ${\boldsymbol\alpha}$. 
The proof is provided in Appendix~\ref{proof:alg2}. 

\textbf{More on the inverse scaling w.r.t. $\boldsymbol{\gamma\left( s_n \right)}$.} This multiplicative factor is likely a consequence of the countable nature of arms (as opposed to finite). When $K=2$, $\alpha_1 \leqslant 1/2$, and the burn-in phase $s_n$ has a fixed duration independent of $n$, the upper bound in Theorem~\ref{thm:alg2} reduces to $\mathcal{O}\left( \beta_{\underline{\Delta},2}^{-1}\left( \log n/\underline{\Delta} + \underline{\Delta}/\alpha_1 \right)\right)$, where the big-Oh only hides absolute constants. Evidently, there is an inflation by $\beta_{\underline{\Delta},2}^{-1}$ relative to the optimal $\mathcal{O}\left( \log n/\underline{\Delta} \right)$ rate achievable in the paradigmatic two-armed bandit with gap $\underline{\Delta}$. By setting $s_n$ as a coercive sub-logarithmic function of the horizon $n$ (e.g., $s_n = \sqrt{\log n}$), one can shave off the $\beta_{\underline{\Delta},2}^{-1}$ factor to achieve $\mathcal{O}\left( \log n/\underline{\Delta} \right)$ regret. This establishes tightness of the instance-dependent lower bound in Theorem~\ref{thm:lower_bound_info} when $K=2$. On the other hand, owing to the dependence of $\gamma\left( s_n \right)$ (and $\beta_{\underline{\Delta},2}$) on $\underline{\Delta}$, the worst-case (instance-independent) upper bound of \texttt{ALG2}$(n)$ can be observed from Theorem~\ref{thm:alg2} to be bounded away from $\Omega\left( \sqrt{n} \right)$. However, recall that Theorem~\ref{thm:alg1} already settles the issue of characterizing the optimal minimax rate when $K=2$ (up to logarithmic factors in $n$). Thus, we provide a complete characterization of the complexity of this problem when $K=2$, thereby answering all the open problems in \cite{kalvit2020finite}. For $K>2$, Theorem~\ref{thm:alg2} guarantees an upper bound of $\mathcal{O}\left( K^3\bar{\Delta}/\delta^2\log n \right)$ under a coercive sub-logarithmic burn-in phase. In this case, characterizing the optimal dependence on $\bar{\Delta}$ and $\delta$ remains an open problem. The scaling w.r.t. $K$, however, cannot be improved to $\mathcal{O}(K)$ as suggested by the $\Omega\left( K\log K \right)$ lower bound in Theorem~\ref{thm:regret-K}. A full characterization of the complexity of the general setting with $K>2$ arm-types remains challenging and is left to future work.

\subsection{Towards fully sequential adaptive strategies: Optimism in exploration}
\label{subsec:UCB}

In this section, we revisit the UCB-based adaptive policy proposed in \cite{kalvit2020finite} for $K=2$. The policy is restated as \texttt{ALG3} below after suitable modifications for reasons discussed next. The original policy (Algorithm~2 in cited paper) achieves an instance-dependent regret of $\mathcal{O}\left( \log n \right)$. Additionally, this algorithm enjoys the benefit of being anytime in $n$ owing to the use of \texttt{UCB1} \citep{auer2002} as a subroutine. However, it suffers a major limitation through its dependence on ex ante knowledge of the support of reward distributions. In particular, the algorithm requires the reward distributions associated with the arm-types to have ``full support'' on $[0,1]$, e.g., only distributions such as Bernoulli$(\cdot)$, Uniform on $[0,1]$, Beta$(\cdot,\cdot)$, etc., are amenable to its performance guarantees; Uniform on $[0,0.5]$, on the other hand, is not. We identify a simple fix to this issue: Drawing inspiration from the design of \texttt{ALG2}$(n)$, we propose adding a centered Gaussian noise term to the difference-of-reward statistic (see step~(6) of Algorithm~\ref{alg:ALG3}) to essentially create an unbounded support. This rids the algorithm of fragility w.r.t. the reward support while also preserving $\mathcal{O}\left( \log n \right)$ regret guarantees (see Theorem~\ref{thm:alg3}).

\textbf{Remark.} The original Algorithm~2 in cited paper uses a threshold that is distinct from $4\sqrt{m\log m}$ (see step~(6) of Algorithm~\ref{alg:ALG3}); the choice of $4\sqrt{m\log m}$ here aims to unify the technical presentation with \texttt{ALG2}$(n)$, and facilitate a more transparent comparison between the corresponding upper bounds.


\begin{algorithm}
\caption{\texttt{ALG3 (Nested UCB1 for $\mathtt{K=2}$ types)}}
\label{alg:ALG3}
\begin{algorithmic}[1]
\BState {\textbf{Initialize new epoch} (resets clock $t\gets 0$)\textbf{:}} Query two \emph{new} arms; call it set $\mathcal{A}=\{1,2\}$.
\State Play each arm in $\mathcal{A}$ once; observe rewards $\left\lbrace X_{a,1} : a\in\mathcal{A} \right\rbrace$.
\State Minimum per-arm sample count $m\gets 1$.
\State Generate a standard Gaussian random variable $\mathcal{Z}$.
\For{$t \in\{3,4,...\}$}
\If {$\left\lvert \mathcal{Z} + \sum_{j=1}^{m}\left( X_{1,j}-X_{2,j} \right) \right\rvert < 4\sqrt{m\log m}$}
\State Permanently discard $\mathcal{A}$ and repeat from step (1).
\Else
\State Play arm~$a_t\in\arg\max_{a\in\mathcal{A}}\left( \frac{\sum_{j=1}^{N_a(t-1)}X_{a,j}}{N_a(t-1)} + \sqrt{\frac{2\log(t-1)}{N_a(t-1)}} \right)$.
\State Observe reward $X_{a_t,N_{a_t}(t)}$.
\If {$m < \min_{a\in\mathcal{A}} N_a(t)$}
\State $m \gets m+1$.
\EndIf
\EndIf
\EndFor
\end{algorithmic}
\end{algorithm}

\textbf{Discussion of Algorithm~\ref{alg:ALG3}.} Similar to \texttt{ALG2}$(n)$, \texttt{ALG3} also has an episodic dynamic with exactly one pair of arms played per episode. The distinction, however, resides in the fact that \texttt{ALG3} plays arms according to \texttt{UCB1} in every episode as opposed to playing them equally often until committing to the empirically superior one. Secondly, unlike \texttt{ALG2}$(n)$, \texttt{ALG3} never ``commits'' to an arm (or a consideration set). The implication is that the algorithm will keep querying new consideration sets throughout the playing horizon; this property is at the core of its anytime nature. Despite these differences, the performance guarantees of the two algorithms are essentially identical when $K=2$, as the next result illustrates. The proof is provided in Appendix~\ref{proof:alg3}.

\begin{theorem}[Upper bound on the regret of \texttt{\textbf{ALG3}} when $\boldsymbol{K=2}$]
\label{thm:alg3}
The expected cumulative regret of the policy $\pi$ given by \texttt{ALG3} after any number of pulls $n \geqslant 2$ is bounded as
\begin{align*}
\mathbb{E}R_n^\pi \leqslant \frac{C}{\beta_{\underline{\Delta},2}}\left( \frac{\log n}{\underline{\Delta}} + \frac{\underline{\Delta}}{\alpha_1} \right),
\end{align*}
where $\beta_{\underline{\Delta},2}$ is as defined in \eqref{eqn:beta_K} with $\delta\gets\underline{\Delta}$ and $K\gets 2$, and $C$ is some absolute constant.
\end{theorem}

\textbf{Remark.} It is possible to shave off the $\beta_{\underline{\Delta},2}$ factor by introducing in \texttt{ALG3} a horizon-dependent burn-in phase \`a la \texttt{ALG2}$(n)$. This may be achieved at the expense of \texttt{ALG3}'s anytime property.

\textbf{Limitation of \texttt{\textbf{ALG3}}.} The performance stated in Theorem~\ref{thm:alg3} together with its anytime property might appear to give an edge to \texttt{ALG3} over \texttt{ALG2}$(n)$. However, the former is theoretically disadvantaged in that its logarithmic upper bound is not currently amenable to extensions to the general $K$-typed setting. The issue traces its roots to the use of \texttt{UCB1} as a subroutine. The concentration behavior of \texttt{UCB1} leveraged towards the analysis of \texttt{ALG3} when $K=2$ fails to hold when $K>2$, rendering proofs intractable. This is illustrated via a simple example with $K=3$ types discussed below.

\textbf{Technical issues with generalizing \texttt{\textbf{ALG3}} to $\boldsymbol{K}$ types.} When $K=2$, there are only two possibilities for what a consideration set could be; arms can have means that are either (i) distinct, or (ii) equal. In the former case, an optimal arm is guaranteed to exist in the consideration set and \texttt{UCB1} will spend the bulk of its sampling effort on it, which is good for regret performance. In the latter scenario, since arms have equal means, \texttt{UCB1} will split samples approximately equally between the two with high probability (see Theorem~4(i) in \cite{kalvit2020finite}); subsequently the consideration set will be discarded within a finite number of samples in expectation (see steps (6) and (7) of \texttt{ALG3}). Contrast this with an alternative setting with $K=3$ and mean rewards $\mu_1 > \mu_2 > \mu_3$. A natural generalization of \texttt{ALG3} (see \texttt{ALG4} in Appendix~\ref{appendix:ALG4}) will query consideration sets of size $3$. Thus, a query can potentially return one arm with mean $\mu_2$ and two with mean $\mu_3$. Since an optimal arm (mean $\mu_1$) is missing, the algorithm will incur linear regret on this set; it is therefore imperative to discard it at the earliest. Unfortunately though, \texttt{UCB1} will invest an overwhelming majority of its sampling effort in the ``locally optimal'' arm (mean $\mu_2$) and allocate logarithmically fewer samples among the other two. This logarithmic rate of sampling arms with mean $\mu_3$ is proof-inhibiting (vis-\`a-vis the $K=2$ case where the rate is linear as previously discussed), making it difficult to theoretically answer if \texttt{ALG4} might still be able to discard the arms within, say, logarithmically many pulls of the horizon. This is an open research question and at the moment, an $\mathcal{O}\left( \log n \right)$ bound exists only for $K=2$; we could only establish asymptotic-optimality ($o(n)$ regret) when $K>2$ (see Theorem~\ref{thm:alg4}). Among other things, identifying the optimal (instance-dependent) scaling factors w.r.t. $\left({\boldsymbol\mu},{\boldsymbol\alpha}\right)$ and the optimal order of minimax regret when $K>2$ remain open problems. 

\section{Concluding remarks and open problems}

This paper provides a first-order characterization of the complexity of the $K$-typed countable-armed bandit problem with matching lower and upper bounds for $K=2$. For $K>2$, we establish an instance-dependent upper bound of $\mathcal{O}\left( K^3\bar{\Delta}/\delta^2\log n \right)$ and show that the scaling w.r.t. $K$ cannot beat $\Omega\left( K\log K \right)$; the latter property differentiates this setting fundamentally from the classical $K$-armed problem. Another key takeaway from our work is that achievable regret in this setting only has a \emph{second-order} dependence on the reservoir distribution, i.e., dependence on $\boldsymbol\alpha$ only manifests through sub-logarithmic terms (see Theorem~\ref{thm:alg2} and \ref{thm:alg3}). Although this work is predicated on countably many arms, our algorithms can easily be adapted to settings with a large but finite number of arms. For example, the result on second-order dependence w.r.t. $\boldsymbol\alpha$ has profound implications for the $N$-armed bandit problem with $K$ arm-types, where each type is characterized by a unique mean reward. A naive implementation of standard MAB algorithms in this setting will result in a regret that scales linearly with $N$. Instead, one can simulate a $K$-typed reservoir over the $N$ arms and deploy \texttt{ALG2}$(n)$ to achieve an $\mathcal{O}\left(K^3\right)$ scaling of the leading term; if $K\ll N$, performance improvement can be substantial vis-\`a-vis naive MAB algorithms. Another important direction concerns adaptivity to $K$: This paper provides algorithms that adapt to $\boldsymbol\alpha$ assuming perfect knowledge of $K$; performance characterization given only an approximation thereof remains an open problem.

\acks{We thank the reviewers for their feedback and suggestions.}

\bibliography{ref}

\begin{thebibliography}{23}
\providecommand{\natexlab}[1]{#1}
\providecommand{\url}[1]{\texttt{#1}}
\expandafter\ifx\csname urlstyle\endcsname\relax
  \providecommand{\doi}[1]{doi: #1}\else
  \providecommand{\doi}{doi: \begingroup \urlstyle{rm}\Url}\fi

\bibitem[Agrawal et~al.(2019)Agrawal, Avadhanula, Goyal, and Zeevi]{mnl-bandit}
Shipra Agrawal, Vashist Avadhanula, Vineet Goyal, and Assaf Zeevi.
\newblock Mnl-bandit: A dynamic learning approach to assortment selection.
\newblock \emph{Operations Research}, 67\penalty0 (5):\penalty0 1453--1485,
  2019.

\bibitem[Auer et~al.(2002)Auer, Cesa-Bianchi, and Fischer]{auer2002}
Peter Auer, Nicolo Cesa-Bianchi, and Paul Fischer.
\newblock Finite-time analysis of the multiarmed bandit problem.
\newblock \emph{Machine learning}, 47\penalty0 (2-3):\penalty0 235--256, 2002.

\bibitem[Banerjee et~al.(2017)Banerjee, Gollapudi, Kollias, and
  Munagala]{banerjee2017segmenting}
Siddhartha Banerjee, Sreenivas Gollapudi, Kostas Kollias, and Kamesh Munagala.
\newblock Segmenting two-sided markets.
\newblock In \emph{Proceedings of the 26th International Conference on World
  Wide Web}, pages 63--72, 2017.

\bibitem[Berry et~al.(1997)Berry, Chen, Zame, Heath, Shepp,
  et~al.]{berry1997bandit}
Donald~A Berry, Robert~W Chen, Alan Zame, David~C Heath, Larry~A Shepp, et~al.
\newblock Bandit problems with infinitely many arms.
\newblock \emph{The Annals of Statistics}, 25\penalty0 (5):\penalty0
  2103--2116, 1997.

\bibitem[Bonald and Proutiere(2013)]{bonald2013two}
Thomas Bonald and Alexandre Proutiere.
\newblock Two-target algorithms for infinite-armed bandits with bernoulli
  rewards.
\newblock In \emph{Advances in Neural Information Processing Systems}, pages
  2184--2192, 2013.

\bibitem[Bui et~al.(2012)Bui, Johari, and Mannor]{bui2012clustered}
Loc Bui, Ramesh Johari, and Shie Mannor.
\newblock Clustered bandits.
\newblock \emph{arXiv preprint arXiv:1206.4169}, 2012.

\bibitem[Carpentier and Valko(2015)]{carpentier2015simple}
Alexandra Carpentier and Michal Valko.
\newblock Simple regret for infinitely many armed bandits.
\newblock In \emph{International Conference on Machine Learning}, pages
  1133--1141, 2015.

\bibitem[Chan and Hu(2018)]{chan2018infinite}
Hock~Peng Chan and Shouri Hu.
\newblock Infinite arms bandit: Optimality via confidence bounds.
\newblock \emph{arXiv preprint arXiv:1805.11793}, 2018.

\bibitem[Chandrasekaran and Karp(2014)]{chandrasekaran2014finding}
Karthekeyan Chandrasekaran and Richard Karp.
\newblock Finding a most biased coin with fewest flips.
\newblock In \emph{Conference on Learning Theory}, pages 394--407. PMLR, 2014.

\bibitem[Chaudhuri and Kalyanakrishnan(2018)]{chaudhuri2018quantile}
Arghya~Roy Chaudhuri and Shivaram Kalyanakrishnan.
\newblock Quantile-regret minimisation in infinitely many-armed bandits.
\newblock In \emph{UAI}, pages 425--434, 2018.

\bibitem[de~Heide et~al.(2021)de~Heide, Cheshire, M{\'e}nard, and
  Carpentier]{de2021bandits}
Rianne de~Heide, James Cheshire, Pierre M{\'e}nard, and Alexandra Carpentier.
\newblock Bandits with many optimal arms.
\newblock In \emph{Advances in Neural Information Processing Systems},
  volume~34, pages 22457--22469, 2021.

\bibitem[Durrett(2019)]{durrett2019probability}
Rick Durrett.
\newblock \emph{Probability: theory and examples}, volume~49.
\newblock Cambridge university press, 2019.

\bibitem[Flajolet et~al.(1992)Flajolet, Gardy, and
  Thimonier]{flajolet1992birthday}
Philippe Flajolet, Daniele Gardy, and Lo{\"y}s Thimonier.
\newblock Birthday paradox, coupon collectors, caching algorithms and
  self-organizing search.
\newblock \emph{Discrete Applied Mathematics}, 39\penalty0 (3):\penalty0
  207--229, 1992.

\bibitem[Hoeffding(1963)]{hoeffding}
W~Hoeffding.
\newblock Probability inequalities for sums of bounded random variables.
\newblock \emph{Journal of the American Statistical Association}, 58\penalty0
  (301):\penalty0 13--30, 1963.

\bibitem[Jamieson et~al.(2016)Jamieson, Haas, and Recht]{jamieson2016detection}
Kevin Jamieson, Daniel Haas, and Ben Recht.
\newblock On the detection of mixture distributions with applications to the
  most biased coin problem.
\newblock \emph{arXiv preprint arXiv:1603.08037}, 2016.

\bibitem[Johari et~al.(2021)Johari, Kamble, and Kanoria]{johari2021matching}
Ramesh Johari, Vijay Kamble, and Yash Kanoria.
\newblock Matching while learning.
\newblock \emph{Operations Research}, 69\penalty0 (2):\penalty0 655--681, 2021.

\bibitem[Kalvit and Zeevi(2020)]{kalvit2020finite}
Anand Kalvit and Assaf Zeevi.
\newblock From finite to countable-armed bandits.
\newblock In \emph{Advances in Neural Information Processing Systems},
  volume~33, pages 8259--8269, 2020.

\bibitem[Kalvit and Zeevi(2021)]{kalvit2021closer}
Anand Kalvit and Assaf Zeevi.
\newblock A closer look at the worst-case behavior of multi-armed bandit
  algorithms.
\newblock In \emph{Advances in Neural Information Processing Systems},
  volume~34, pages 8807--8819, 2021.

\bibitem[Lai and Robbins(1985)]{lai}
Tze~Leung Lai and Herbert Robbins.
\newblock Asymptotically efficient adaptive allocation rules.
\newblock \emph{Advances in applied mathematics}, 6\penalty0 (1):\penalty0
  4--22, 1985.

\bibitem[Lattimore and Szepesv{\'a}ri(2020)]{lattimore2020bandit}
Tor Lattimore and Csaba Szepesv{\'a}ri.
\newblock \emph{Bandit algorithms}.
\newblock Cambridge University Press, 2020.

\bibitem[Slivkins et~al.(2019)]{slivkins2019introduction}
Aleksandrs Slivkins et~al.
\newblock Introduction to multi-armed bandits.
\newblock \emph{Foundations and Trends{\textregistered} in Machine Learning},
  12\penalty0 (1-2):\penalty0 1--286, 2019.

\bibitem[Wang et~al.(2009)Wang, Audibert, and Munos]{wang2009}
Yizao Wang, Jean-Yves Audibert, and R{\'e}mi Munos.
\newblock Algorithms for infinitely many-armed bandits.
\newblock In \emph{Advances in Neural Information Processing Systems}, pages
  1729--1736, 2009.

\bibitem[Zhu and Nowak(2020)]{zhu2020regret}
Yinglun Zhu and Robert Nowak.
\newblock On regret with multiple best arms.
\newblock \emph{Advances in Neural Information Processing Systems},
  33:\penalty0 9050--9060, 2020.

\end{thebibliography}

\appendix

\section*{Supplementary material: General organization}

\begin{enumerate}

\item Appendix~\ref{sec:experiments} provides numerical experiments.

\item Appendix~\ref{appendix:ALG4} generalizes \texttt{ALG3} to $K>2$ and states performance guarantees thereof.

\item Appendix~\ref{appendix:lower_bound_info} provides the proof of Theorem~\ref{thm:lower_bound_info}.

\item Appendix~\ref{appendix:lower_bound0} provides the proof of Theorem~\ref{thm:lower_bound0}.

\item Appendix~\ref{proof:regret-K} provides the proof of Theorem~\ref{thm:regret-K}.

\item Appendix~\ref{proof:alg1} provides the proof of Theorem~\ref{thm:alg1}.

\item Appendix~\ref{appendix:propK} provides the proof of Proposition~\ref{propK}.

\item Appendix~\ref{proof:alg2} provides the proof of Theorem~\ref{thm:alg2}.

\item Appendix~\ref{appendix:aux_alg3} provides auxiliary results used in the analysis of \texttt{ALG3}.

\item Appendix~\ref{proof:alg3} provides the proof of Theorem~\ref{thm:alg3}.

\item Appendix~\ref{appendix:aux_alg4} provides auxiliary results used in the analysis of \texttt{ALG4}.

\item Appendix~\ref{proof:alg4} provides the proof of Theorem~\ref{thm:alg4}.

\end{enumerate}

\section{Numerical experiments}
\label{sec:experiments}

We evaluate the empirical performance of our algorithms for $K=2$ and $K=3$ on synthetic data.

\textbf{Experiments.} In what follows, the graphs show the performance of different algorithms simulated on synthetic data. The horizon is capped at $n = 10^5$ for $K=2$ and at $n=10^4$ for $K=3$. Each regret trajectory is averaged over at least $100$ independent experiments (sample-paths). The shaded regions indicate standard $95\%$ confidence intervals. For horizon-dependent algorithms, regret is plotted for discrete values of the horizon $n$ indicated by ``$\ast$'' and interpolated; for anytime algorithms, regret accrued until each $t\in\{1,...,n\}$ is plotted. 

\textbf{Baseline policies.} We will benchmark the performance of our algorithms against two policies: (i) \texttt{Sampling-UCB} \citep{de2021bandits}, and (ii) \texttt{ETC-}$\infty(2)$ \citep{kalvit2020finite}. The former is a UCB-styled policy based on front-loading exploration of \emph{new} arms (Theorem~\ref{thm:lower_bound0} thus applies to this policy). It is, however, noteworthy that \texttt{Sampling-UCB} is predicated on ex ante knowledge of (a lower bound on) the probability $\alpha_1$ of sampling an optimal arm from the reservoir; we reemphasize that this is not the setting of interest in our paper. Furthermore, its regret scales as $\tilde{\mathcal{O}}\left( \log n/\left(\alpha_1\underline{\Delta}\right) \right)$ (up to poly-logarithmic factors in $1/\underline{\Delta}$), which is inferior in terms of its dependence on $\alpha_1$ relative to \texttt{ALG2}$(n)$ and \texttt{ALG3} (see Theorem~\ref{thm:alg2} and \ref{thm:alg3} respectively). There exist other algorithms as well (see, e.g., \cite{chaudhuri2018quantile, zhu2020regret}) developed for formulations with prohibitively large number of arms. However, these are either sensitive to certain parametric assumptions on the probability of sampling an optimal arm, or focus on a different notion of regret altogether; both directions remain outside the ambit of our setting.

The second policy \texttt{ETC-}$\infty(2)$ is a non-adaptive explore-then-commit-styled algorithm for reservoirs with $K=2$ types; this policy requires ex ante knowledge of a lower bound on the difference between the two mean rewards. Although \texttt{ETC-}$\infty(2)$ was originally proposed only for $K=2$, it is easily generalizable and we present in Algorithm~\ref{alg:ETC-bad} a version (\texttt{ETC-}$\infty(K)$) that is adapted to $K$ types.

\begin{figure}[H]
\begin{minipage}[t]{0.48\linewidth}
\centering
\includegraphics[width=\textwidth]{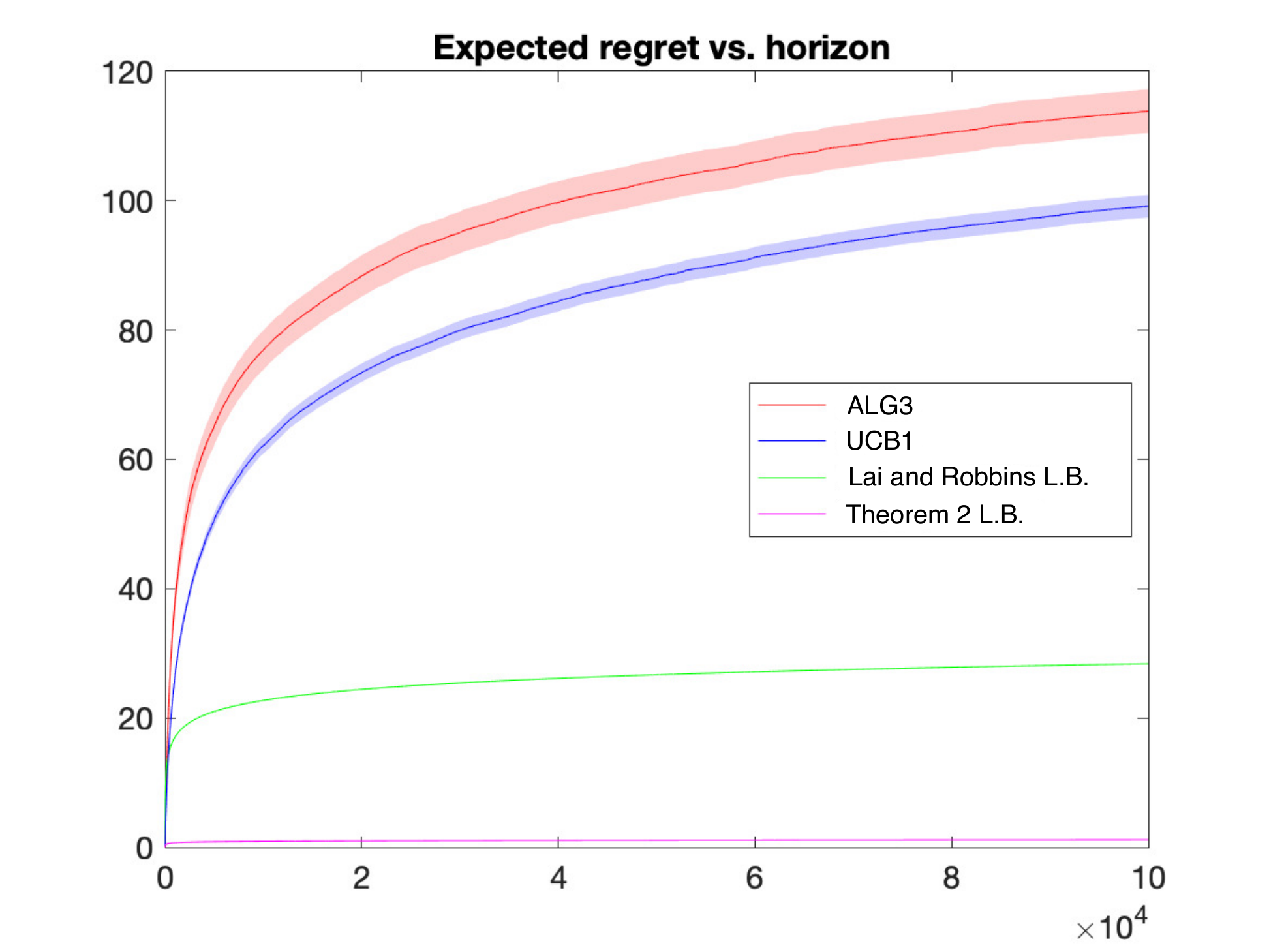}
\caption{$\boldsymbol{K=2}$ \textbf{and} $\boldsymbol{{\boldsymbol\alpha}=\left(1/2,1/2 \right)}$: Achievable regret in 2-CAB vis-\`a-vis 2-MAB.}
\label{fig:figure3}
\end{minipage}
\hfill
\begin{minipage}[t]{0.48\linewidth}
\centering
\includegraphics[width=\textwidth]{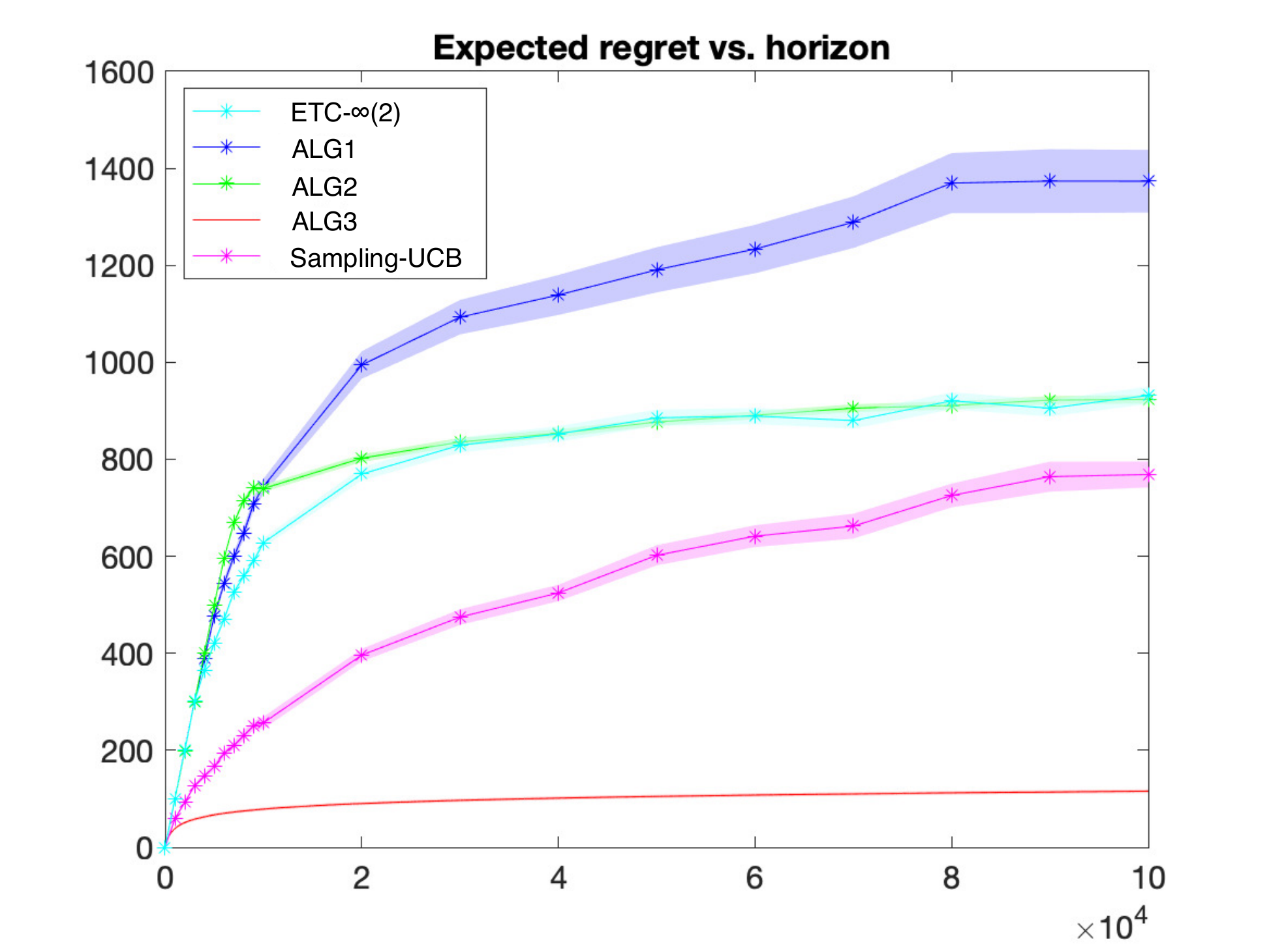}
\caption{$\boldsymbol{K=2}$ \textbf{and} $\boldsymbol{{\boldsymbol\alpha}=\left(1/2,1/2 \right)}$: An instance with Bernoulli $0.6, 0.4$ rewards.}
\label{fig:figure4}
\end{minipage}
\end{figure}

\begin{figure}[H]
\begin{minipage}[t]{0.48\linewidth}
\centering
\includegraphics[width=\textwidth]{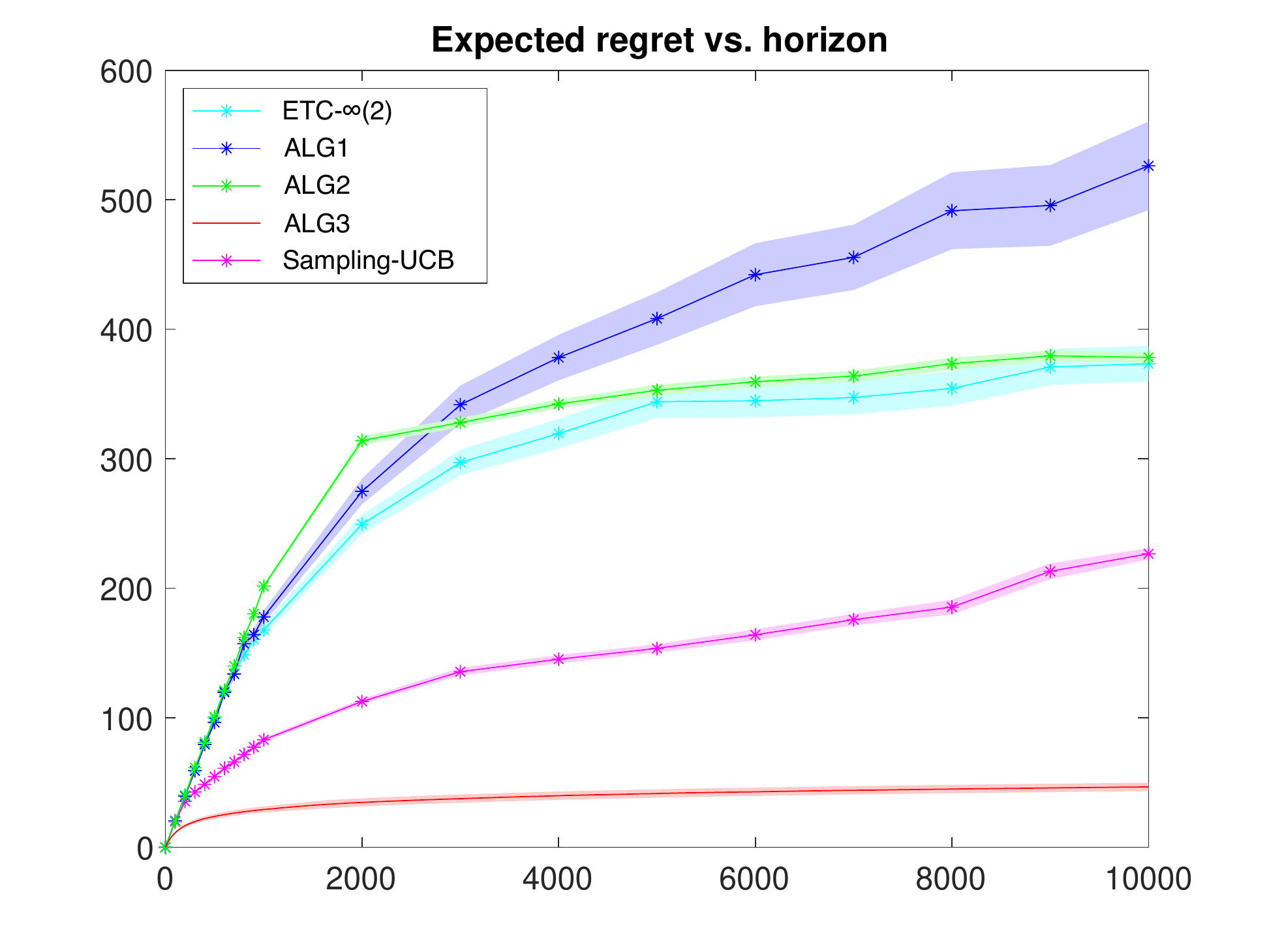}
\caption{$\boldsymbol{K=2}$ \textbf{and} $\boldsymbol{{\boldsymbol\alpha}=\left(1/2,1/2 \right)}$: An instance with Bernoulli $0.9, 0.5$ rewards.}
\label{fig:figure1}
\end{minipage}
\hfill
\begin{minipage}[t]{0.48\linewidth}
\centering
\includegraphics[width=\textwidth]{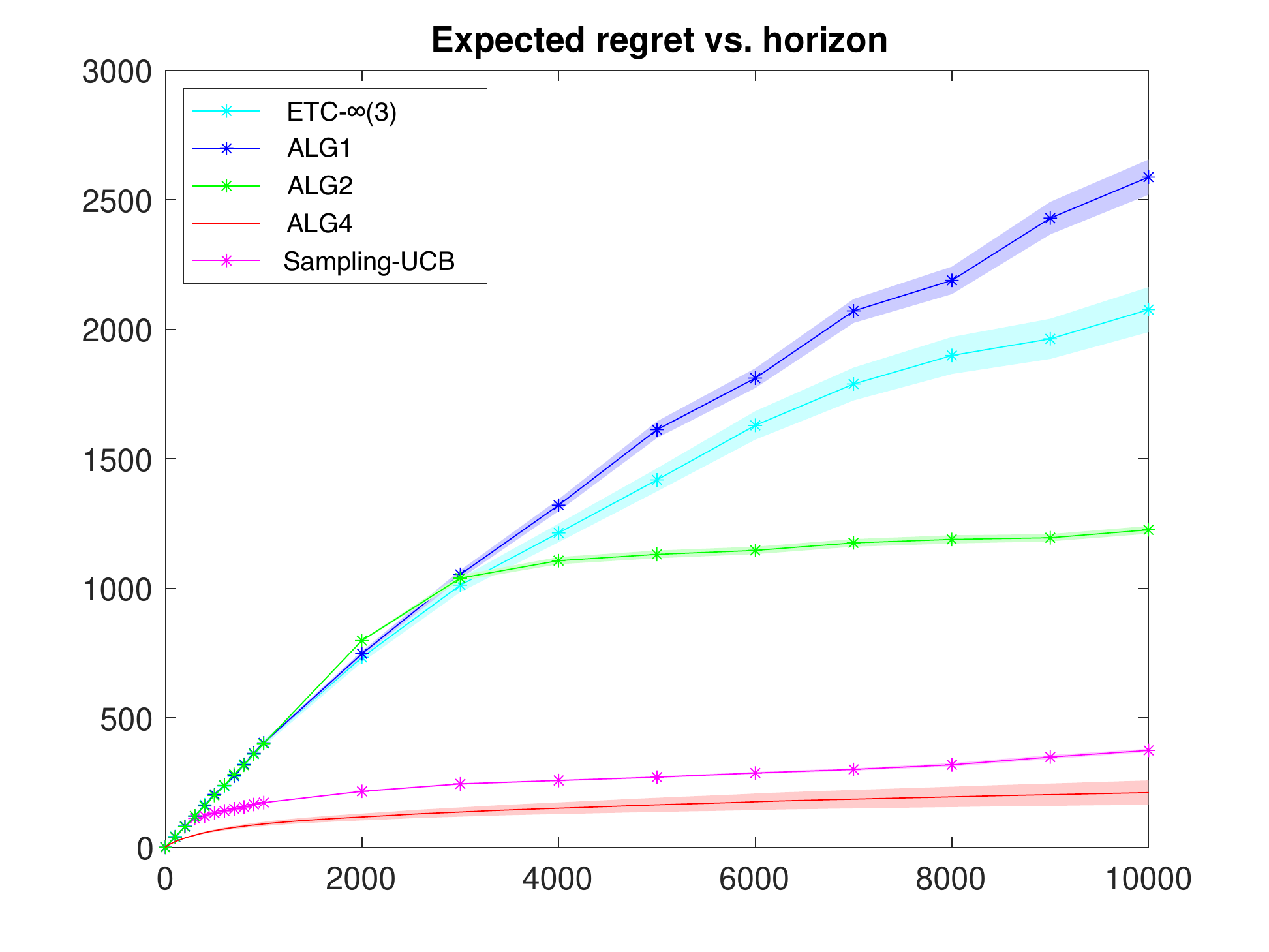}
\caption{$\boldsymbol{K=3}$ \textbf{and} $\boldsymbol{{\boldsymbol\alpha}=\left( 1/3,1/3,1/3 \right)}$: Instance with Bernoulli $0.9, 0.5, 0.1$ rewards.}
\label{fig:figure2}
\end{minipage}
\end{figure}

\textbf{Setup 1 [Figures~\ref{fig:figure3}, \ref{fig:figure4} and \ref{fig:figure1}].} In this setting, we consider $K=2$ with $\alpha_1=0.5$, i.e., two equiprobable arm-types, characterized by Bernoulli$(0.6)$ and Bernoulli$(0.4)$ rewards. Via this setup, we intend to illustrate the difference between the empirical performance achievable in the countable-armed setting vis-\`a-vis its traditional two-armed counterpart. Refer to Figure~\ref{fig:figure3}. The red curve indicates the empirical performance of \texttt{ALG3} in this setting. For reference, the blue one shows the empirical performance of \texttt{UCB1} \citep{auer2002} in a two-armed bandit with Bernoulli(0.6) and Bernoulli(0.4) rewards; the green curve indicates the best achievable instance-dependent regret \citep{lai} in said two-armed configuration. As expected, the regret of \texttt{ALG3} is inflated relative to \texttt{UCB1}. This is owing to the $\beta_{\delta,2} \leqslant 1$ factor present in the denominator of \texttt{ALG3}'s upper bound; characterization of the sharpest lower bound on the probability in \eqref{eqn:beta_K} (see Proposition~\ref{propK}) is challenging owing to the limited theoretical tools available to this end and we leave it as an open problem at the moment. Figure~\ref{fig:figure4} shows the empirical performance of the algorithms proposed in this paper as well as \texttt{Sampling-UCB} initialized with $\alpha_1 = 1/2$ and \texttt{ETC-}$\infty(2)$ initialized with $\underline{\delta} = \delta/2=0.1$. Evidently, the (adaptive) explore-then-commit approach in \texttt{ALG2}$(n)$ outperforms the pre-specified exploration schedule-based approach of \texttt{ALG1}$(n)$, and performs almost as good as the \emph{gap-aware} approach in \texttt{ETC-}$\infty(2)$. While \texttt{Sampling-UCB} outmatches all explore-then-commit styled approaches, the best performing algorithm is \texttt{ALG3}. Surprisingly, this is despite the fact that the theoretical performance bounds for \texttt{ALG2}$(n)$ and \texttt{ALG3} are identical (modulo numerical multiplicative constants) when $K=2$ and $\alpha_1 \leqslant 0.5$ (see Theorem~\ref{thm:alg2} and \ref{thm:alg3}). A similar hierarchy in performances is also observable in Figure~\ref{fig:figure1}, which corresponds to a slightly ``easier'' instance with $\delta = 0.4$ (as opposed to $0.2$) and equiprobable Bernoulli$(0.9)$ and Bernoulli$(0.5)$ rewards.

\textbf{Setup 2 [Figure~\ref{fig:figure2}].} Here, we consider a setting with $K=3$ arm-types characterized by Bernoulli rewards with means $0.9, 0.5, 0.1$, each occurring with probability $1/3$. We compare the performance of \texttt{ALG1}$(n)$, \texttt{ALG2}$(n)$ and \texttt{ALG4} (generalization of \texttt{ALG3} to $K \geqslant 2$; see Appendix~\ref{appendix:ALG4}) with \texttt{ETC-}$\infty(3)$ initialized with $\underline{\delta}=\delta/2 = 0.2$, and \texttt{Sampling-UCB} initialized with $\alpha_1 = 1/3$. It is noteworthy that despite \texttt{ALG4}'s significantly superior empirical performance relative to aforementioned algorithms, only a weak $o(n)$ theoretical guarantee on its regret is currently available (see Theorem~\ref{thm:alg4}) due to reasons discussed earlier in the paper. Investigating best achievable rates under \texttt{ALG4} is an area of active research at the moment.

\section{More algorithms for reservoirs with $\boldsymbol{K\geqslant 2}$ arm-types}
\label{appendix:ALG4}

\begin{algorithm}[H]
\caption{\texttt{ALG4 (Nested UCB1 for $\mathtt{K}$ types)}}
\label{alg:ALG4}
\begin{algorithmic}[1]
\BState {\textbf{Initialize new epoch} (resets clock $t\gets 0$)\textbf{:}} Query $K$ \emph{new} arms; call it set $\mathcal{A}=\{1,...,K\}$.
\State Play each arm in $\mathcal{A}$ once; observe rewards $\left\lbrace X_{a,1} : a\in\mathcal{A} \right\rbrace$.
\State Minimum per-arm sample count $m\gets 1$.
\State Generate ${K \choose 2}$ independent standard Gaussian random variables $\left\lbrace \mathcal{Z}_{a,b} : a,b\in\mathcal{A}, a<b \right\rbrace$.
\For{$t \in\{K+1,K+2,...\}$}
\If {$\exists$ $a,b\in\mathcal{A}$, $a<b$ s.t. $\left\lvert \mathcal{Z}_{a,b} + \sum_{j=1}^{m}\left( X_{a,j}-X_{b,j} \right) \right\rvert < 4\sqrt{m\log m}$}
\State Permanently discard $\mathcal{A}$ and repeat from step (1).
\Else
\State Play arm~$a_t\in\arg\max_{a\in\mathcal{A}}\left( \frac{\sum_{j=1}^{N_a(t-1)}X_{a,j}}{N_a(t-1)} + \sqrt{\frac{2\log(t-1)}{N_a(t-1)}} \right)$.
\State Observe reward $X_{a_t,N_{a_t}(t)}$.
\If {$m < \min_{a\in\mathcal{A}} N_a(t)$}
\State $m \gets m+1$.
\EndIf
\EndIf
\EndFor
\end{algorithmic}
\end{algorithm}

\begin{theorem}[Upper bound on the regret of \texttt{\textbf{ALG4}}]
\label{thm:alg4}
The expected cumulative regret of the policy $\pi$ given by \texttt{ALG4} after any number $n \geqslant 1$ of plays is bounded as
\begin{align*}
\mathbb{E}R_n^\pi \leqslant \frac{CK}{\beta_{\delta,K}}\left( \frac{\log n}{\underline{\Delta}} + \bar{\Delta} \right) + o\left( \frac{\bar{\Delta}n}{\beta_{\delta,K}\prod_{i=1}^{K}\alpha_i} \right),
\end{align*} 
where $C$ is some absolute constant, $\beta_{\delta,K}$ is as defined in \eqref{eqn:beta_K}, and the little-Oh is asymptotic in $n$ and only hides multiplicative factors in $K$.
\end{theorem}

Proof is provided in Appendix~\ref{proof:alg4}.

\begin{algorithm}
\caption{\texttt{ETC-$\infty$(K)}}
\label{alg:ETC-bad}
\begin{algorithmic}[1]
\BState \textbf{Input:} (i) Horizon of play $n$, (ii) A lower bound $\underline{\delta}\in\left(0,\delta\right]$ on the minimal reward gap $\delta$. 
\State Set budget $T=n$.
\BState \textbf{Initialize new epoch:} Query $K$ \emph{new} arms; call it consideration set $\mathcal{A}=\{1,...,K\}$.
\State Set exploration duration $L = \left\lceil 2\underline{\delta}^{-2}\log n \right\rceil$.
\State $m \gets \min \left( L, \left\lfloor T/K \right\rfloor \right)$.
\State Play each arm in $\mathcal{A}$ $m$ times; observe rewards $\left\lbrace \left( X_{1,j}, ..., X_{K,j}\right) : j=1,...,m \right\rbrace$.
\State Update budget: $T \gets T - Km$.
\If {$\exists$ $a,b\in\mathcal{A}$, $a<b$ s.t. $\left\lvert {\sum_{j=1}^{m}(X_{a,j}-X_{b,j})} \right\rvert < \underline{\delta} m$}
\State Permanently discard $\mathcal{A}$, and repeat from step (3).
\Else
\State Permanently commit to arm $a^{\ast} \in \arg\max_{a\in\mathcal{A}}\left\lbrace \sum_{j=1}^{m}X_{a,j} \right\rbrace$.
\EndIf
\end{algorithmic}
\end{algorithm}

\section{Proof of Theorem~\ref{thm:lower_bound_info}}
\label{appendix:lower_bound_info}


\textbf{Notation.} For each $i\in\{1,2\}$, let $\mathcal{G}_i(x)$ be an arbitrary collection of distributions with mean $x\in\mathbb{R}$. The tuple $\left( \mathcal{G}_1(x), \mathcal{G}_2(y) \right)$ will be referred to as an \emph{instance}.

Since the horizon of play is fixed at $n$, the decision maker may play at most $n$ distinct arms. Therefore, it suffices to focus only on the sequence of the first $n$ arms that may be played. A \textit{realization} of an instance $\nu = \left( \mathcal{G}_1(\mu_1), \mathcal{G}_2(\mu_2) \right)$ is defined as the $n$-tuple $r\equiv \left( r_i \right)_{1 \leqslant i \leqslant n}$, where $r_i\in\mathcal{G}_1(\mu_1)\cup\mathcal{G}_2(\mu_2)$ denotes the reward distribution of arm $i\in\{1,...,n\}$. It must be noted that the decision maker need not play every arm in $r$. Let $i^\ast := \arg\max_{i\in\{1,2\}}\mu_i$. Suppose that the distribution over possible realizations of $\nu = \left( \mathcal{G}_1(\mu_1), \mathcal{G}_2(\mu_2) \right)$ in $\left\lbrace r : r_i\in\mathcal{G}_1(\mu_1)\cup\mathcal{G}_2(\mu_2),\ 1 \leqslant i \leqslant n \right\rbrace$ satisfies $\mathbb{P}\left(r_i \in \mathcal{G}_{i^\ast}\left(\mu_{i^\ast}\right)\right)=\alpha^\ast$ (where $\alpha^\ast\in(0,1)$ is arbitrary) for all $i\in\{1,...,n\}$, i.e., optimal arms occur in the reservoir with probability $\alpha^\ast$.

Recall that the cumulative pseudo-regret after $n$ plays of a policy $\pi$ on $\nu = \left( \mathcal{G}_1(\mu_1), \mathcal{G}_2(\mu_2) \right)$ is given by $R_n^{\pi}(\nu)=\sum_{t=1}^{n}\left(\mu_{i^\ast} - \mu_{\mathcal{T}\left(\pi_t\right)}\right)$, where $\mathcal{T}(\pi_t)\in\{1,2\}$ indicates the type of the arm played by $\pi$ at time $t$. Our goal is to lower bound $\mathbb{E}R_n^{\pi}(\nu)$, where the expectation is w.r.t. the randomness in $\pi$ as well as the distribution over possible realizations of $\nu$. To this end, we define the notion of expected cumulative regret of $\pi$ on a realization $r$ of $\nu = \left( \mathcal{G}_1(\mu_1), \mathcal{G}_2(\mu_2) \right)$ by
\begin{align*}
S_n^{\pi}(\nu,r) := \mathbb{E}^{\pi}\left[\sum_{t=1}^{n}\left( \mu_{i^\ast} - \mu_{\mathcal{T}\left(\pi_t\right)}\right)\right],
\end{align*}
where the expectation $\mathbb{E}^{\pi}$ is w.r.t. the randomness in $\pi$. Note that $\mathbb{E}R_n^{\pi}(\nu) = \mathbb{E}^{\nu}S_n^{\pi}(\nu,r)$, where the expectation $\mathbb{E}^{\nu}$ is w.r.t. the distribution over possible realizations of $\nu$. We define our problem class $\mathcal{N}_{\underline{\Delta}}$ as the collection of $\underline{\Delta}$-separated instances given by
\begin{align*}
\mathcal{N}_{\underline{\Delta}} := \left\lbrace \left( \mathcal{G}_1(\mu_1), \mathcal{G}_2(\mu_2) \right) : \mu_1 - \mu_2 = \underline{\Delta},\ (\mu_1,\mu_2)\in\mathbb{R}^2 \right\rbrace.
\end{align*}

Fix an arbitrary $\underline{\Delta}>0$ and consider an instance $\nu=\left( \{Q_1\}, \{Q_2\} \right)\in\mathcal{N}_{\underline{\Delta}}$, where $(Q_1,Q_2)$ are unit-variance Gaussian distributions with means $(\mu_1,\mu_2)$ respectively. Consider an arbitrary realization $r\in\{Q_1,Q_2\}^n$ of $\nu$ and let $\mathcal{I}\subseteq\{1,...,n\}$ denote the set of inferior arms in $r$ (arms with reward distribution $Q_2$). Consider another instance $\nu^{\prime}\in\mathcal{N}_{\underline{\Delta}}$ given by $\nu^{\prime}=\left(\{\widetilde{Q}_1\}, \{Q_1\}\right)$, where $\widetilde{Q}_1$ is another unit variance Gaussian with mean $\mu_1+\underline{\Delta}$. Now consider a realization $r^{\prime}\in\{\widetilde{Q}_1,Q_1\}^n$ of $\nu^{\prime}$ that is such that the arms at positions in $\mathcal{I}$ have distribution $\widetilde{Q}_1$ while those at positions in $\{1,...,n\}\backslash\mathcal{I}$ have distribution $Q_1$. Notice that $\mathcal{I}$ is the set of optimal arms in $r^{\prime}$ (arms with reward distribution $\widetilde{Q}_1$). Then, the following always holds:
\begin{align*}
S_n^{\pi}(\nu,r) + S_n^{\pi}(\nu^{\prime},r^{\prime}) \geqslant \left(\frac{\underline{\Delta} n}{2}\right)\left( \mathbb{P}^{\pi}_{\nu,r}\left( \sum_{i\in\mathcal{I}}N_i(n) > \frac{n}{2} \right) + \mathbb{P}^{\pi}_{\nu^{\prime},r^\prime}\left( \sum_{i\in\mathcal{I}}N_i(n) \leqslant \frac{n}{2} \right) \right),
\end{align*}
where $\mathbb{P}^{\pi}_{\nu,r}(\cdot)$ and $\mathbb{P}^{\pi}_{\nu^\prime,r^\prime}(\cdot)$ denote the probability measures w.r.t. the instance-realization pairs $(\nu,r)$ and $(\nu^{\prime},r^{\prime})$ respectively, and $N_i(n)$ denotes the number of plays up to and including time $n$ of arm $i\in\{1,...,n\}$. Using the Bretagnolle-Huber inequality (Theorem~14.2 of \cite{lattimore2020bandit}), we obtain
\begin{align*}
S_n^{\pi}(\nu,r) + S_n^{\pi}(\nu^{\prime},r^{\prime}) \geqslant \left(\frac{\underline{\Delta} n}{4}\right)\exp\left( -\text{D}\left(\mathbb{P}^{\pi}_{\nu,r},\mathbb{P}^{\pi}_{\nu^{\prime},r^{\prime}} \right) \right),
\end{align*}
where $\text{D}\left(\mathbb{P}^{\pi}_{\nu,r},\mathbb{P}^{\pi}_{\nu^{\prime},r^{\prime}} \right)$ denotes the KL-Divergence between $\mathbb{P}^{\pi}_{\nu,r}$ and $\mathbb{P}^{\pi}_{\nu^{\prime},r^{\prime}}$. Using Divergence decomposition (Lemma~15.1 of \cite{lattimore2020bandit}), we further obtain
\begin{align*}
S_n^{\pi}(\nu,r) + S_n^{\pi}(\nu^{\prime},r^{\prime}) &\geqslant \left(\frac{\underline{\Delta} n}{4}\right)\exp\left( - \left(\frac{\text{D}\left( Q_2,\widetilde{Q}_{1} \right)}{\underline{\Delta}}\right)S_n^{\pi}(\nu,r) \right) = \left(\frac{\underline{\Delta} n}{4}\right)\exp\left( -2\underline{\Delta} S_n^{\pi}(\nu,r) \right),
\end{align*}
where the equality follows since $\widetilde{Q}_1$ and $Q_2$ are unit variance Gaussian distributions with means separated by $2\underline{\Delta}$. Next, taking the expectation $\mathbb{E}^{\nu}$ on both sides followed by a direct application of Jensen's inequality yields
\begin{align}
\mathbb{E}R_n^{\pi}(\nu) + \mathbb{E}^{\nu}S_n^{\pi}(\nu^{\prime},r^{\prime}) \geqslant \left(\frac{\underline{\Delta} n}{4}\right)\exp\left( -2\underline{\Delta} \mathbb{E}R_n^{\pi}(\nu) \right). \label{eqn:temp1}
\end{align}

Consider the $\mathbb{E}^{\nu}S_n^{\pi}(\nu^{\prime},r^{\prime})$ term in \eqref{eqn:temp1}. Using a simple change-of-measure argument, we obtain
\begin{align}
\mathbb{E}^{\nu}S_n^{\pi}(\nu^{\prime},r^{\prime}) = \mathbb{E}^{{\nu}^{\prime}}\left[ S_n^{\pi}(\nu^{\prime},r^{\prime}) \left( \frac{1-\alpha^\ast}{\alpha^\ast} \right)^{2\left(\Lambda(r^{\prime})-n/2\right)} \right], \notag
\end{align}
where $\Lambda(r^{\prime})$ is the number of optimal arms in realization $r^\prime$.

Since $\alpha^\ast$ is arbitrary, we fix $\alpha^\ast=1/2$ to obtain
\begin{align}
\mathbb{E}^{\nu}S_n^{\pi}(\nu^{\prime},r^{\prime}) = \mathbb{E}^{{\nu}^{\prime}}S_n^{\pi}(\nu^{\prime},r^{\prime}) = \mathbb{E}R_n^{\pi}\left(\nu^\prime\right), \label{eqn:temp2}
\end{align}

Now, from \eqref{eqn:temp1} and \eqref{eqn:temp2}, we have that for $\alpha^\ast=1/2$,
\begin{align}
&\mathbb{E}R_n^\pi\left( \nu \right) + \mathbb{E}R_n^\pi\left( \nu^\prime \right) \geqslant \frac{\underline{\Delta} n}{4}\exp\left( -2\underline{\Delta}\mathbb{E}R_n^\pi\left( \nu \right) \right) \notag \\
\implies\ &\tilde{R}_n \geqslant \frac{\underline{\Delta} n}{8}\exp\left( -2\underline{\Delta}\tilde{R}_n \right), \label{eqn:critical_bound}
\end{align}
where $\tilde{R}_n := \max\left( \mathbb{E}R_n^\pi\left( \nu \right), \mathbb{E}R_n^\pi\left( \nu^\prime \right) \right)$.

\noindent\textbf{Instance-dependent lower bound}

The assertion of the theorem follows from the fact that the inequality \eqref{eqn:critical_bound} is fulfilled only if for any $\varepsilon\in(0,1)$, $\tilde{R}_n$ satisfies for all $n$ large enough $\tilde{R}_n \geqslant \left( 1-\varepsilon \right)\log n/\left( 2\underline{\Delta} \right)$. Therefore, there exists an instance $\nu$ with gap $\underline{\Delta}$ such that $\mathbb{E}R_n^\pi\left( \nu \right) \geqslant C\log n/\underline{\Delta}$ for some absolute constant $C$ and $n$ large enough, whenever $\alpha^\ast = 1/2$. In fact, said statement holds for all $\alpha^\ast \leqslant 1/2$ since the policy $\pi$ satisfies Definition~\ref{def:consistency}.

\noindent\textbf{Instance-independent (minimax) lower bound}

Since $\tilde{R}_n \leqslant \underline{\Delta} n$, it follows from \eqref{eqn:critical_bound} that
\begin{align*}
\tilde{R}_n \geqslant \frac{\underline{\Delta} n}{8}\exp\left( -2\underline{\Delta}^2 n \right).
\end{align*}
Setting $\underline{\Delta} = 1/\sqrt{n}$, and noting that the inequality, in fact, holds for all $\alpha^\ast \leqslant 1/2$ (owing to the admissibility of $\pi$; see Definition~\ref{def:consistency}), proves the stated assertion. \hfill $\square$

\section{Proof of Theorem~\ref{thm:lower_bound0}}
\label{appendix:lower_bound0}

Note that this result is stated for general $K \geqslant 2$ and is not specific to $K=2$. In fact, the nature of the set of possible sub-optimal types is inconsequential to the proof that follows as long as said set is at least $\underline{\Delta}$-separated from the optimal mean reward. Consider an arbitrary policy $\pi\in\tilde{\Pi}$. Denote by $A_n^\pi$ the number of \emph{distinct} arms played by $\pi$ until time $n$. Consider an arbitrary $k\in\{1,...,n\}$. Then, conditioned on $A_n^\pi = k$, the expected cumulative regret incurred by $\pi$ is at least 
\begin{align}
\mathbb{E}\left[ R_n^\pi | A_n^\pi = k \right] &\geqslant (1-\alpha_1)\underline{\Delta} k + (1-\alpha_1)^k\underline{\Delta}(n-k) =: f(k). \label{eqn:oracle-LB}
\end{align}

\textbf{Intuition behind \eqref{eqn:oracle-LB}.} Each of the $k$ arms played during the horizon has at least one pull associated with it. Consider a clairvoyant policy coupled to $\pi$ that learns the best among the $A_n^\pi$ arms played by $\pi$ as soon as each has been pulled exactly once, i.e., after a total of $A_n^\pi$ pulls. Clearly, the regret incurred by said clairvoyant policy lower bounds $\mathbb{E}R_n^\pi$. Further, since $A_n^\pi$ is independent of the sample-history of arms, it follows that the $A_n^\pi$ arms are statistically identical. Thus, conditioned on $A_n^\pi=k$, the expected regret from the first $k$ pulls of the clairvoyant policy is at least $(1-\alpha_1)\underline{\Delta} k$. Also, the probability that each of the $k$ arms is inferior-typed is $(1-\alpha_1)^k$; the clairvoyant policy thus incurs a regret of at least $(1-\alpha_1)^k\underline{\Delta}(n-k)$ going forward. This explains the lower bound in \eqref{eqn:oracle-LB}. Therefore, for any $k\in\{1,2,...,n\}$, we have
\begin{align*}
\mathbb{E}\left[ R_n^\pi | A_n^\pi = k \right] &\geqslant \min_{k\in\{1,2,...,n\}}f(k) \geqslant \min_{x\in[0,n]}f(x).
\end{align*}

We will show that $f(x)$ is strictly convex over $[0,n]$ with $f^\prime(0)<0$ and $f^\prime(n)>0$. Then, it would follow that $f(\cdot)$ admits a unique minimizer $x_n^\ast\in(0,n)$ given by the solution to $f^\prime(x)=0$. The minimum $f\left( x_n^\ast \right)$ will turn out to be logarithmic in $n$. Observe that
\begin{align*}
f^\prime(x) &= (1-\alpha_1)\underline{\Delta} + (1-\alpha_1)^x\underline{\Delta}\left[ (n-x) \log(1-\alpha_1) - 1 \right], \\
f^{\prime\prime}(x) &= -(1-\alpha_1)^x\underline{\Delta}\left[ 2 - (n-x)\log(1-\alpha_1) \right]\log(1-\alpha_1).
\end{align*}

Since $\underline{\Delta} > 0$, it follows that $f^{\prime\prime}(x) > 0$ over $[0,n]$. Further, note that
\begin{align*}
f^\prime(0) &= -\alpha_1\underline{\Delta} + \underline{\Delta} n \log(1-\alpha_1) < 0, \\
f^\prime(n) &= (1-\alpha_1)\underline{\Delta} - (1-\alpha_1)^n \underline{\Delta} > 0.
\end{align*}

Solving $f^\prime\left( x_n^\ast \right) = 0$ for the unique minimizer $x_n^\ast$, we obtain
\begin{align*}
&\left( \frac{1}{1-\alpha_1} \right)^{x_n^\ast-1} - 1 = \left( n - x_n^\ast \right)\log\left( \frac{1}{1-\alpha_1} \right) \\
\implies &\left( \frac{1}{1-\alpha_1} \right)^{x_n^\ast} + x_n^\ast\log\left( \frac{1}{1-\alpha_1} \right) > n\log\left( \frac{1}{1-\alpha_1} \right) \\
\implies &2\left( \frac{1}{1-\alpha_1} \right)^{x_n^\ast} > n\log\left( \frac{1}{1-\alpha_1} \right),
\end{align*}
where the last inequality follows using $y > \log y$. Therefore, we have
\begin{align*}
&\left( \frac{1}{1-\alpha_1} \right)^{x_n^\ast} > \frac{n}{2}\log\left( \frac{1}{1-\alpha_1} \right) \\
\implies &x_n^\ast > \frac{\log n + \log\log\left( \frac{1}{1-\alpha_1} \right) - \log 2}{\log\left( \frac{1}{1-\alpha_1} \right)}.
\end{align*}

Thus, for any $k\in\{1,...,n\}$,
\begin{align*}
&\mathbb{E}\left[ R_n^\pi | A_n^\pi = k \right] \geqslant f\left( x_n^\ast \right) > (1-\alpha_1)\underline{\Delta} x_n^\ast > (1-\alpha_1) \left( \frac{\log n + \log\log\left( \frac{1}{1-\alpha_1} \right) - \log 2}{\log\left( \frac{1}{1-\alpha_1} \right)} \right)\underline{\Delta} \\
\implies &\mathbb{E} R_n^\pi \geqslant (1-\alpha_1) \left( \frac{\log n + \log\log\left( \frac{1}{1-\alpha_1} \right) - \log 2}{\log\left( \frac{1}{1-\alpha_1} \right)} \right)\underline{\Delta} \\
\implies &\inf_{\pi\in\tilde{\Pi}}\frac{\mathbb{E} R_n^\pi}{\log n} \geqslant (1-\alpha_1) \left( \frac{1}{\log\left( \frac{1}{1-\alpha_1} \right)} + \frac{\log\log\left( \frac{1}{1-\alpha_1} \right) - \log 2}{\left(\log n\right)\log\left( \frac{1}{1-\alpha_1} \right)} \right)\underline{\Delta} \\
\implies &\inf_{\pi\in\tilde{\Pi}}\frac{\mathbb{E} R_n^\pi}{\log n} \underset{\mathrm{(\dag)}}{\geqslant} (1-\alpha_1) \left( \frac{1-\alpha_1}{\alpha_1} + \frac{\log\log\left( \frac{1}{1-\alpha_1} \right) - \log 2}{\left(\log n\right)\log\left( \frac{1}{1-\alpha_1} \right)} \right)\underline{\Delta} \\
\implies &\inf_{\pi\in\tilde{\Pi}}\frac{\mathbb{E} R_n^\pi}{\log n} \geqslant \frac{(1-\alpha_1)^2\underline{\Delta}}{\alpha_1} + (1-\alpha_1)\left(\frac{\log\log\left( \frac{1}{1-\alpha_1} \right) - \log 2}{\left(\log n\right)\log\left( \frac{1}{1-\alpha_1} \right)} \right)\underline{\Delta},
\end{align*}
where $(\dag)$ follows using $\log y \leqslant y-1$. Taking the appropriate limit now proves the assertion. \hfill $\square$

\section{Proof of Theorem~\ref{thm:regret-K}}
\label{proof:regret-K}

The reservoir distribution is given by $\boldsymbol\alpha = \left( \alpha_1, ..., \alpha_K \right)$. In the full information setting, the decision maker observes the true mean reward of an arm immediately upon pulling it. Let $\pi = \left( \pi_t : t = 1,2,... \right)$ be the policy that pulls a new arm from the reservoir in each period. Let $N$ denote the first time at which one arm of each of the $K$ types is collected under $\pi$. Then, it follows from classical results (see Theorem~4.1 in \cite{flajolet1992birthday}) for the \emph{Coupon-collector problem} that
\begin{align}
\mathbb{E}N = \int_{0}^{\infty} \left( 1 - \prod_{j=1}^{K}\left( 1- \exp\left(-\alpha_j y \right) \right) \right) dy &\underset{\mathrm{(\dag)}}{\geqslant} \int_{0}^{\infty} \left( 1 - \prod_{j=1}^{K}\left( 1- \exp\left(-\frac{y}{K}\right) \right) \right) dy \notag \\
&\underset{\mathrm{(\ddag)}}{=} \sum_{j=1}^{K}\frac{K}{j} \notag \\
&\geqslant K\log K, \label{eqn:EN}
\end{align}
where $(\dag)$ follows as $\prod_{j=1}^{K}\left( 1- \exp\left(-\alpha_j y \right) \right)$ is maximized when $\boldsymbol\alpha$ is the Uniform distribution; $(\ddag)$ is a classical result (see previous reference). The optimal policy $\pi^\ast$ follows $\pi$ until time $N$, and subsequently commits to the arm with the highest mean among the first $N$ arms. The lifetime regret of $\pi^\ast$ is then given by
\begin{align}
\mathbb{E}R_\infty^{\pi^\ast} &= \mathbb{E}\left[ \sum_{t=1}^{N} \sum_{i=2}^{K}\left( \mu_1 - \mu_i \right)\mathbbm{1}\left\lbrace \mathcal{T}\left( \pi_t \right) = i\right\rbrace \right] \notag \\
&= \mathbb{E}\left[ \sum_{t=1}^{\infty} \sum_{i=2}^{K}\left( \mu_1 - \mu_i \right)\mathbbm{1}\left\lbrace \mathcal{T}\left( \pi_t \right) = i,\ t \leqslant N \right\rbrace \right] \notag \\
&= \sum_{t=1}^{\infty} \sum_{i=2}^{K}\left( \mu_1 - \mu_i \right)\mathbb{P}\left( \mathcal{T}\left( \pi_t \right) = i,\ t \leqslant N \right), \label{eqn:Regret-pi-star}
\end{align}
where the last equality follows from Tonelli's Theorem. Note that
\begin{align}
\mathbb{P}\left( \mathcal{T}\left( \pi_t \right) = i,\ t \leqslant N \right) &= \mathbb{P}\left( \mathcal{T}\left( \pi_t \right) = i \right) - \mathbb{P}\left( \mathcal{T}\left( \pi_t \right) = i,\ t > N\right) \notag \\
&= \alpha_i - \mathbb{P}\left( \left. \mathcal{T}\left( \pi_t \right) = i\ \right\rvert\ t > N\right)\mathbb{P}\left( t > N \right) \notag \\
&\underset{\mathrm{(\star)}}{=}\alpha_i - \mathbb{P}\left( \mathcal{T}\left( \pi_t \right) = i \right)\mathbb{P}\left( t > N \right) \notag \\
&= \alpha_i - \alpha_i\mathbb{P}\left( t > N \right) \notag \\
&= \alpha_i \mathbb{P}\left( N \geqslant t \right), \label{eqn:prod-prob}
\end{align}
where $(\star)$ follows since $\mathcal{T}\left( \pi_t \right)$ is i.i.d. in time $t$. Therefore, from \eqref{eqn:Regret-pi-star} and \eqref{eqn:prod-prob}, we have
\begin{align}
\mathbb{E}R_\infty^{\pi^\ast} &= \sum_{t=1}^{\infty} \sum_{i=2}^{K}\alpha_i\left( \mu_1 - \mu_i \right)\mathbb{P}\left( N \geqslant t \right) \notag \\
&= \sum_{i=2}^{K}\alpha_i\left( \mu_1 - \mu_i \right)\sum_{t=1}^{\infty}\mathbb{P}\left( N \geqslant t \right) \notag \\
&= \sum_{i=2}^{K}\alpha_i\left( \mu_1 - \mu_i \right)\mathbb{E}N. \label{eqn:Regret-pi-star-final}
\end{align}
Finally, from \eqref{eqn:Regret-pi-star-final} and \eqref{eqn:EN}, one obtains
\begin{align}
\mathbb{E}R_\infty^{\pi^\ast} &\geqslant \sum_{i=2}^{K}\alpha_i\left( \mu_1 - \mu_i \right)K\log K. \notag
\end{align} \hfill $\square$

\section{Proof of Theorem~\ref{thm:alg1}}
\label{proof:alg1}

Let $\left\lbrace \left( X_{1,j}^l, ..., X_{K,j}^l \right) : j = 1,...,m_l \right\rbrace$ be the reward sequence associated with the $K$ arms played in the $l$\textsuperscript{th} epoch, where $m_l = \left\lceil e^{2\sqrt{l}} \log n \right\rceil$. Let $\mathcal{A} := \{1,...,K\}$ and define
\begin{align*}
I := \inf \left\lbrace l \in \mathbb{N} : \left\lvert \sum_{j=1}^{m_l} \left( X_{a,j}^l - X_{b,j}^{l} \right) \right\rvert \geqslant 2m_le^{-\sqrt{l}}\ \forall\ a,b\in\mathcal{A}, a<b \right\rbrace.
\end{align*}
Then,
\begin{align}
\mathbb{P}\left( I \geqslant k \right) &= \mathbb{P}\left( \bigcap_{l=1}^{k-1} \bigcup_{a,b\in\mathcal{A}, a<b} \left\lbrace \left\lvert \sum_{j=1}^{m_l} \left( X_{a,j}^l - X_{b,j}^{l} \right) \right\rvert < 2m_le^{-\sqrt{l}} \right\rbrace \right) \notag \\
&= \prod_{l=1}^{k-1}\mathbb{P}\left( \bigcup_{a,b\in\mathcal{A}, a<b} \left\lbrace \left\lvert \sum_{j=1}^{m_l} \left( X_{a,j}^l - X_{b,j}^l \right) \right\rvert < 2m_le^{-\sqrt{l}} \right\rbrace \right) \notag \\
&\leqslant \prod_{l=1}^{k-1}\left[ 1 - \mathbb{P}\left( \texttt{D} \right) + \mathbb{P}\left( \texttt{D} \right) \mathbb{P}_{\texttt{D}}\left( \bigcup_{a,b\in\mathcal{A}, a<b} \left\lbrace \left\lvert \sum_{j=1}^{m_l} \left( X_{a,j}^l - X_{b,j}^l \right) \right\rvert < 2m_le^{-\sqrt{l}} \right\rbrace \right) \right], \notag
\end{align}
where \texttt{D} denotes the event that the $K$ arms played in epoch $l$ are ``all-distinct,'' i.e., no two arms belong to the same type, $\mathbb{P}_\mathtt{D}(\cdot) := \mathbb{P}\left(\cdot|\mathtt{D}\right)$ denotes the corresponding conditional measure, and $\mathbb{P}\left( \texttt{D} \right) = K!\prod_{i=1}^{K}\alpha_i$. Using the Union bound, we obtain
\begin{align}
\mathbb{P}\left( I \geqslant k \right) &\leqslant \prod_{l=1}^{k-1}\left[ 1 - \mathbb{P}\left( \texttt{D} \right) + \mathbb{P}\left( \texttt{D} \right)\sum_{a,b\in\mathcal{A}, a<b}\mathbb{P}_{\texttt{D}}\left( \left\lvert \sum_{j=1}^{m_l} \left( X_{a,j}^l - X_{b,j}^l \right) \right\rvert < 2m_le^{-\sqrt{l}} \right) \right]. \label{eqn:idhar_dekho_firse}
\end{align}

Define $\tau := K\sum_{l=1}^{I}m_l$. Consider the following events:
\begin{align*}
\mathtt{E}_1 &:= \left\lbrace \text{None of the arms played in epoch $I$ belongs to the optimal type} \right\rbrace, \\
\mathtt{E}_2 &:= \left\lbrace \text{At least one optimal-typed arm is played in epoch $I$, and the empirically best arm is \underline{not} optimal-typed} \right\rbrace.
\end{align*}

Recall that $\bar{\Delta} = \mu_1 - \mu_K$ denotes the maximal sub-optimality gap. Then, the cumulative pseudo-regret $R_n$ (superscript $\pi$ suppressed for notational convenience) of \texttt{ALG1}$(n)$ is bounded as
\begin{align*}
R_n &\leqslant \mathbbm{1}\left\lbrace \tau \leqslant n \right\rbrace \left[ \bar{\Delta}\tau + \mathbbm{1}\left\lbrace \mathtt{E}_1\cup\mathtt{E}_2 \right\rbrace \bar{\Delta} n \right] + \mathbbm{1}\left\lbrace \tau > n \right\rbrace \bar{\Delta} n \\
&\leqslant \bar{\Delta}\tau + \left( \mathbbm{1}\left\lbrace \tau \leqslant n, \mathtt{E}_1 \right\rbrace + \mathbbm{1} \left\lbrace \tau \leqslant n, \mathtt{E}_2 \right\rbrace \right)\bar{\Delta} n + \mathbbm{1}\left\lbrace \tau > n \right\rbrace \bar{\Delta} n.
\end{align*} 

Taking expectations,
\begin{align*}
\mathbb{E}R_n &\leqslant \bar{\Delta}\mathbb{E}\tau + \left[ \mathbb{P}\left( \tau \leqslant n, \mathtt{E}_1 \right) + \mathbb{P} \left( \tau \leqslant n, \mathtt{E}_2 \right) \right]\bar{\Delta} n + \mathbb{P}\left( \tau > n \right) \bar{\Delta} n \\
&\leqslant 2\bar{\Delta}\mathbb{E}\tau + \left[ \mathbb{P}\left( \tau \leqslant n, \mathtt{E}_1 \right) + \mathbb{P} \left( \tau \leqslant n, \mathtt{E}_2 \right) \right]\bar{\Delta} n,
\end{align*}
where the last step uses Markov's inequality. Therefore,
\begin{align*}
\mathbb{E}R_n &\leqslant 2K\bar{\Delta}\mathbb{E}\left[ I m_I \right] + \left[ \mathbb{P}\left( \tau \leqslant n, \mathtt{E}_1 \right) + \mathbb{P} \left( \tau \leqslant n, \mathtt{E}_2 \right) \right]\bar{\Delta} n \\
&\leqslant 4K\bar{\Delta}\mathbb{E}\left[ I e^{2\sqrt{I}} \right]\log n + \left[ \mathbb{P}\left( \tau \leqslant n, \mathtt{E}_1 \right) + \mathbb{P} \left( \tau \leqslant n, \mathtt{E}_2 \right) \right]\bar{\Delta} n.
\end{align*}

\subsection{Upper bounding $\mathbb{E}\left[ I e^{2\sqrt{I}}\right]$}

Recall that $\delta = \min_{1 \leqslant i < j \leqslant K}\left( \mu_i - \mu_j \right)$ denotes the smallest gap between any two distinct mean rewards. Then, on the event \texttt{D}, for any $a,b\in\mathcal{A}, a<b$, we either have $\mathbb{E}\left[ X_{a,j}^l - X_{b,j}^l \right] \geqslant \delta$ or $\mathbb{E}\left[ X_{a,j}^l - X_{b,j}^l \right] \leqslant -\delta$. Without loss of generality, suppose that $\mathbb{E}\left[ X_{a,j}^l - X_{b,j}^l \right] \geqslant \delta$. Then,
\begin{align*}
\mathbb{P}_\mathtt{D}\left( \left\lvert \sum_{j=1}^{m_l} \left( X_{a,j}^l - X_{b,j}^l \right) \right\rvert < 2m_le^{-\sqrt{l}} \right) &\leqslant \mathbb{P}_\mathtt{D}\left( \sum_{j=1}^{m_l} \left( X_{a,j}^l - X_{b,j}^l \right) < 2m_le^{-\sqrt{l}} \right) \\
&= \mathbb{P}_\mathtt{D}\left( \sum_{j=1}^{m_l} \left( X_{a,j}^l - X_{b,j}^l -\delta \right) < -m_l\left( \delta - 2e^{-\sqrt{l}} \right) \right).
\end{align*}

Then, for $l > \left\lceil \log^2\left( 4/\delta \right) \right\rceil =: k^\ast$, one has that
\begin{align}
\mathbb{P}_\mathtt{D}\left( \left\lvert \sum_{j=1}^{m_l} \left( X_{a,j}^l - X_{b,j}^l \right) \right\rvert < 2m_le^{-\sqrt{l}} \right) \leqslant \mathbb{P}_\mathtt{D}\left( \sum_{j=1}^{m_l} \left( X_{a,j}^l - X_{b,j}^l -\delta \right) < -2m_le^{-\sqrt{l}} \right) \leqslant n^{-2}, \label{eqn:idhar_wala_final}
\end{align}
where the final inequality follows using the Chernoff-Hoeffding bound \citep{hoeffding}, together with the fact that $-1 \leqslant X_{a,j}^l - X_{b,j}^l \leqslant 1$ and $m_l = \left\lceil e^{2\sqrt{l}} \log n \right\rceil$. Using \eqref{eqn:idhar_dekho_firse} and \eqref{eqn:idhar_wala_final}, we obtain for $k > k^\ast+1$ that
\begin{align*}
\mathbb{P}\left( I \geqslant k \right) &\leqslant \prod_{l=1}^{k-1}\left[ 1 - \mathbb{P}\left( \texttt{D} \right) + \mathbb{P}\left( \texttt{D} \right) \sum_{a,b\in\mathcal{A}, a<b} \mathbb{P}_{\texttt{D}}\left( \left\lvert \sum_{j=1}^{m_l} \left( X_{a,j}^l - X_{b,j}^l \right) \right\rvert < 2m_le^{-\sqrt{l}} \right) \right] \\
&\leqslant \prod_{k^\ast< l \leqslant k-1} \left[ 1 - \mathbb{P}\left( \texttt{D} \right) + \frac{K^2\mathbb{P}\left( \texttt{D} \right)}{n^2}\right] \\
&\leqslant \left[ 1 - \frac{3\mathbb{P}\left( \texttt{D} \right)}{4} \right]^{k-k^\ast-1},
\end{align*}
where the last inequality holds for $n\geqslant 2K$.\footnote{We will ensure that all guarantees hold for $n \geqslant K$ by offsetting regret by $2K\bar{\Delta}$ in the end.} Thus, for any $k \geqslant 1$ and $n \geqslant 2K$, we have
\begin{align*}
\mathbb{P}\left( I \geqslant k^\ast + k \right) \leqslant \left[ 1 - \frac{3\mathbb{P}\left( \texttt{D} \right)}{4} \right]^{k-1}.
\end{align*}

Now, 
\begin{align*}
\mathbb{E}\left[ I e^{2\sqrt{I}} \right] &\leqslant k^\ast e^{2\sqrt{k^\ast}} + \sum_{k=1}^{\infty}\left( k^\ast+k\right)e^{2\sqrt{k^\ast+k}}\mathbb{P}\left( I = k^\ast + k \right) \\
&\underset{\mathrm{(\dag)}}{\leqslant} k^\ast e^{2\sqrt{k^\ast}} + \sum_{k=1}^{\infty}\left( k^\ast+k\right)e^{2\sqrt{k^\ast}}e^{2\sqrt{k}}\mathbb{P}\left( I = k^\ast + k \right) \\
&\underset{\mathrm{(\ddag)}}{\leqslant} k^\ast e^{2\sqrt{k^\ast}} + 2\sum_{k=1}^{\infty}k^\ast ke^{2\sqrt{k^\ast}}e^{2\sqrt{k}}\mathbb{P}\left( I = k^\ast + k \right) \\
&\leqslant k^\ast e^{2\sqrt{k^\ast}} + 2\sum_{k=1}^{\infty} k^\ast k e^{2\sqrt{k^\ast}}e^{2\sqrt{k}}\mathbb{P}\left( I \geqslant k^\ast + k \right) \\
&\leqslant k^\ast e^{2\sqrt{k^\ast}} + 2\sum_{k=1}^{\infty} k^\ast ke^{2\sqrt{k^\ast}}e^{2\sqrt{k}}\left[ 1 - \frac{3\mathbb{P}\left( \texttt{D} \right)}{4} \right]^{k-1} \\
&= k^\ast e^{2\sqrt{k^\ast}} \left[ 1 + 2\sum_{k=1}^{\infty}k e^{2\sqrt{k}}\left[ 1 - \frac{3\mathbb{P}\left( \texttt{D} \right)}{4} \right]^{k-1} \right] \\
&\leqslant 4k^\ast e^{2\sqrt{k^\ast}} \sum_{k=1}^{\infty}k e^{2\sqrt{k}}\left[ 1 - \frac{3\mathbb{P}\left( \texttt{D} \right)}{4} \right]^{k-1},
\end{align*}
where $(\dag)$ follows using $\sqrt{k^\ast + k} \leqslant \sqrt{k^\ast} + \sqrt{k}$, and $(\ddag)$ using $k^\ast + k \leqslant 2k^\ast k$ (since $k^\ast, k\in\mathbb{N}$). Define
\begin{align*}
C_{{\boldsymbol\alpha}} := \sum_{k=1}^{\infty}k e^{2\sqrt{k}}\left[ 1 - \frac{3\mathbb{P}\left( \texttt{D} \right)}{4} \right]^{k-1}.
\end{align*}
Since $\mathbb{P}\left( \texttt{D} \right) = K!\prod_{i=1}^{K}\alpha_i$, note that the infinite summation is finite since ${\boldsymbol\alpha} = \left( \alpha_i : i = 1,...,K \right)$ is coordinate-wise bounded away from $0$. Therefore,
\begin{align*}
\mathbb{E}\left[ I e^{2\sqrt{I}} \right] \leqslant 4C_{{\boldsymbol\alpha}} k^\ast e^{2\sqrt{k^\ast}}.
\end{align*}
Note that
\begin{align*}
e^{2\sqrt{k^\ast}} = e^{2\sqrt{\left\lceil \log^2\left(4/\delta\right)\right\rceil}} \leqslant e^{2\sqrt{\log^2\left(4/\delta\right)+1}} \leqslant e^{2\left(\sqrt{\log^2\left(4/\delta\right)}+1\right)} \leqslant \frac{16e^2}{\delta^2}.
\end{align*}
Therefore, in conclusion,
\begin{align*}
\mathbb{E}\left[ I e^{2\sqrt{I}} \right] \leqslant \frac{128e^2C_{{\boldsymbol\alpha}}}{\delta^2}{\log^2\left( \frac{4}{\delta}\right)}.
\end{align*}

\subsection{Upper bounding $\mathbb{P}\left( \tau \leqslant n, \mathtt{E}_1 \right)$}

Note that on the event $\left\lbrace \tau \leqslant n \right\rbrace$, the duration of epoch $I$ is $Km_I$. On the event $\mathtt{E}_1$, the consideration set contains at least two arms that belong to the same type. Without loss of generality, suppose that these arms are indexed $1$ and $2$. Then,
\begin{align*}
\mathbb{P}\left( \tau \leqslant n, \mathtt{E}_1 \right) &\leqslant \mathbb{P}\left( \left\lvert \sum_{j=1}^{m_I}\left( X_{1,j}^I - X_{2,j}^I \right) \right\rvert \geqslant 2m_Ie^{-\sqrt{I}}, \tau \leqslant n \right) \\
&\leqslant \mathbb{P}\left( \left\lvert \sum_{j=1}^{m_I}\left( X_{1,j}^I - X_{2,j}^I \right) \right\rvert \geqslant 2m_Ie^{-\sqrt{I}}, I \leqslant n \right) \\
&\leqslant \sum_{l=1}^{n}\mathbb{P}\left( \left\lvert \sum_{j=1}^{m_l}\left( X_{1,j}^l - X_{2,j}^l \right) \right\rvert \geqslant 2m_le^{-\sqrt{l}} \right) \\
&\leqslant \sum_{l=1}^{n}\frac{2}{n^2} \\
&= \frac{2}{n},
\end{align*}
where the last inequality follows using the Chernoff-Hoeffding bound \citep{hoeffding}.

\subsection{Upper bounding $\mathbb{P}\left( \tau \leqslant n, \mathtt{E}_2 \right)$}

Note that on the event $\left\lbrace \tau \leqslant n \right\rbrace$, the duration of epoch $I$ is $Km_I$. On the event $\mathtt{E}_2$, the consideration set $\mathcal{A}$ contains at least one arm of the optimal type, and the empirically best arm belongs to an inferior type. Without loss of generality, suppose that arm~$1$ belongs to the optimal type and $\mathcal{I} \subset \mathcal{A}$ denotes the set of inferior-typed arms. We then have
\begin{align*}
\mathbb{P}\left( \tau \leqslant n, \mathtt{E}_2 \right) &\leqslant \mathbb{P}\left( \bigcup_{a\in\mathcal{I}} \left\lbrace \sum_{j=1}^{m_I}\left( X_{a,j}^I - X_{1,j}^I \right) \geqslant 2m_Ie^{-\sqrt{I}}, \tau \leqslant n \right\rbrace \right) \\
&\leqslant \sum_{a\in\mathcal{I}}\mathbb{P}\left( \sum_{j=1}^{m_I}\left( X_{a,j}^I - X_{1,j}^I \right) \geqslant 2m_Ie^{-\sqrt{I}}, I \leqslant n \right) \\
&\leqslant \sum_{a\in\mathcal{I}}\sum_{l=1}^{n}\mathbb{P}\left( \sum_{j=1}^{m_l}\left( X_{a,j}^l - X_{1,j}^l \right) \geqslant 2m_le^{-\sqrt{l}} \right) \\
&\leqslant \sum_{a\in\mathcal{I}}\sum_{l=1}^{n}\frac{1}{n^2} \\
&\leqslant \frac{K}{n},
\end{align*}
where the second-to-last inequality follows using the Chernoff-Hoeffding bound \citep{hoeffding} since $\mathbb{E}\left[ X_{a,j}^{l} - X_{1,j}^{l} \right] \leqslant -\underline{\Delta} < 0\ \forall\ a\in\mathcal{I}$.

\subsection{Putting everything together}

In conclusion, the expected cumulative regret of the policy $\pi$ given by \texttt{ALG1}$(n)$ is bounded for any $n\geqslant K$ as
\begin{align*}
\mathbb{E}R_n^{\pi} \leqslant \frac{\tilde{C}_{{\boldsymbol\alpha}}K\bar{\Delta}\log n}{\delta^2}\log^2\left( \frac{4}{\delta} \right) + 4K\bar{\Delta},
\end{align*}
where $\tilde{C}_{{\boldsymbol\alpha}}$ is a finite constant that depends only on ${\boldsymbol\alpha}=\left( \alpha_i : i = 1,...,K \right)$. In particular, $\tilde{C}_{{\boldsymbol\alpha}}$ is given by the following infinite summation:
\begin{align}
\tilde{C}_{{\boldsymbol\alpha}} := 512e^2C_{{\boldsymbol\alpha}} = 512e^2\sum_{k=1}^{\infty}k e^{2\sqrt{k}}\left[ 1 - \frac{3K!\prod_{i=1}^{K}\alpha_i}{4} \right]^{k-1}. \label{eqn:C_alpha_K}
\end{align} \hfill $\square$

\section{Proof of Proposition~\ref{propK}}
\label{appendix:propK}

Consider the following stopping time:
\begin{align*}
\tau := \inf \left\lbrace m \in\mathbb{N} : \exists a,b\in\mathcal{A}, a<b\ \text{s.t.}\ \mathcal{Z}_{a,b} + \sum_{j=1}^{m}\left( X_{a,j} - X_{b,j} \right) < 4\sqrt{m\log m} \right\rbrace.
\end{align*}
Since $\mathbb{P}\left( \bigcap_{m \geqslant 1}\bigcap_{a,b\in\mathcal{A}, a<b}\left\lbrace \left\lvert \mathcal{Z}_{a,b} + \sum_{j=1}^{m}\left( X_{a,j} - X_{b,j} \right) \right\rvert \geqslant 4\sqrt{m\log m} \right\rbrace \right) \geqslant \mathbb{P}(\tau = \infty)$, it suffices to show that $\mathbb{P}(\tau=\infty)$ is bounded away from $0$. To this end, define the following entities:
\begin{align*}
&\Lambda_K := \inf \left\lbrace p \in\mathbb{N} : \sum_{m=p}^{\infty}\frac{1}{m^8} \leqslant \frac{1}{2K^2} \right\rbrace, \\
&T_0 := \max\left( \left\lceil \left( \frac{64}{\delta^2} \right)\log^2\left( \frac{64}{\delta^2} \right) \right\rceil, \Lambda_K \right), \\
&f(x) := x + 4\sqrt{x\log x} & \mbox{for $x\geqslant 1$}.
\end{align*}

\begin{lemma}
\label{lemma:prop1K}
For any $a,b\in\mathcal{A}$ s.t. $a<b$, it is the case that
\begin{align*}
\left\lbrace \mathcal{Z}_{a,b} > f\left( T_0 \right) \right\rbrace \subseteq \bigcap_{m=1}^{T_0} \left\lbrace \mathcal{Z} + \sum_{j=1}^{m}\left( X_{a,j} - X_{b,j} \right) \geqslant 4\sqrt{m\log m} \right\rbrace.
\end{align*}
\end{lemma}

\textbf{Proof of Lemma~\ref{lemma:prop1K}.} Note that 
\begin{align*}
\mathcal{Z}_{a,b} &> f\left( T_0 \right) \\
&= T_0 + 4\sqrt{T_0\log T_0} \\
&\geqslant m + 4\sqrt{m\log m}\ \forall\ 1 \leqslant m \leqslant T_0 \\
&\underset{\mathrm{(\mathfrak{a})}}{\geqslant} \sum_{j=1}^{m}\left( X_{b,j} - X_{a,j} \right) + 4\sqrt{m\log m}\ \forall\ 1 \leqslant m \leqslant T_0 \\
\implies \mathcal{Z}_{a,b} + \sum_{j=1}^{m}\left( X_{a,j} - X_{b,j} \right) &\geqslant 4\sqrt{m\log m}\ \forall\ 1 \leqslant m \leqslant T_0,
\end{align*}
where $(\mathfrak{a})$ follows since the rewards are bounded in $[0,1]$, i.e., $\left\lvert X_{a,j} - X_{b,j} \right\rvert \leqslant 1$. \hfill $\square$

\begin{lemma} 
\label{lemma:prop2K}
For $m \geqslant T_0$, it is the case that
\begin{align*}
\delta \geqslant 8\sqrt{\frac{\log m}{m}}.
\end{align*}
\end{lemma}

\textbf{Proof of Lemma~\ref{lemma:prop2K}.} First of all, note that $T_0 \geqslant \left( {64}/{\delta^2} \right)\log^2\left( {64}/{\delta^2} \right) \geqslant 64$ (since $\delta \leqslant 1$). For $s = \left( {64}/{\delta^2} \right)\log^2\left( {64}/{\delta^2} \right)$, one has 
\begin{align*}
\delta^2 = \frac{64\log^2\left( \frac{64}{\delta^2}\right)}{s} \underset{\mathrm{(\mathfrak{b})}}{\geqslant} \frac{64\left[\log\left( \frac{64}{\delta^2}\right) + 2\log\log\left( \frac{64}{\delta^2} \right) \right]}{s} = \frac{64\log s}{s},
\end{align*}
where $(\mathfrak{b})$ follows since the function $g(x) := x^2 - x - 2\log x$ is monotone increasing for $x \geqslant \log 64$ (think of $\log\left( {64}/{\delta^2}\right)$ as $x$), and therefore attains its minimum at $x = \log 64$; one can verify that this minimum is strictly positive. Furthermore, since $\log s/s$ is monotone decreasing for $s \geqslant 64$, it follows that for any $m \geqslant T_0$, 
\begin{align*}
\delta^2 \geqslant \frac{64\log m}{m}.
\end{align*} \hfill $\square$

Now coming back to the proof of Proposition~\ref{propK}, consider an arbitrary $l\in\mathbb{N}$ such that $l > T_0$. Then,
\begin{align*}
\mathbb{P}\left(\tau \leqslant l \right) &= \mathbb{P}\left( \tau \leqslant l, \mathcal{Z} > f\left(T_0\right) \right) + \mathbb{P}\left( \tau \leqslant l, \mathcal{Z} \leqslant f\left(T_0\right) \right) \\
&\leqslant \mathbb{P}\left( \tau \leqslant l, \mathcal{Z} > f\left(T_0\right) \right) + \Phi\left( f\left(T_0\right) \right).
\end{align*}

Now,
\begin{align*}
&\mathbb{P}\left( \tau \leqslant l, \mathcal{Z} > f\left(T_0\right) \right) \\
=\ &\mathbb{P}\left( \bigcup_{m=1}^{l}\bigcup_{a,b\in\mathcal{A}, a<b}\left\lbrace \mathcal{Z}_{a,b} + \sum_{j=1}^{m}\left( X_{a,j} - X_{b,j} \right) < 4\sqrt{m\log m}, \mathcal{Z}_{a,b} > f\left(T_0\right) \right\rbrace \right) \\
\underset{\mathrm{(\dag)}}{=}\ &\mathbb{P}\left( \bigcup_{m = T_0}^{l}\bigcup_{a,b\in\mathcal{A}, a<b}\left\lbrace \mathcal{Z}_{a,b} + \sum_{j=1}^{m}\left( X_{a,j} - X_{b,j} \right) < 4\sqrt{m\log m}, \mathcal{Z}_{a,b} > f\left(T_0\right) \right\rbrace \right) \\
\leqslant\ &\sum_{m=T_0}^{l} \sum_{a,b\in\mathcal{A}, a<b}\mathbb{P}\left( \mathcal{Z}_{a,b} + \sum_{j=1}^{m}\left( X_{a,j} - X_{b,j} \right) < 4\sqrt{m\log m}, \mathcal{Z}_{a,b} > f\left(T_0\right) \right) \\
=\ &\sum_{m=T_0}^{l} \sum_{a,b\in\mathcal{A}, a<b} \mathbb{P}\left( \mathcal{Z}_{a,b} + \sum_{j=1}^{m}\left( X_{a,j} - X_{b,j} -\delta \right) < -m\left( \delta - 4\sqrt{\frac{\log m}{m}} \right), \mathcal{Z}_{a,b} > f\left(T_0\right) \right) \\
\leqslant\ &\sum_{m=T_0}^{l} \sum_{a,b\in\mathcal{A}, a<b} \mathbb{P}\left( \sum_{j=1}^{m}\left( X_{a,j} - X_{b,j} -\delta \right) < -m\left( \delta - 4\sqrt{\frac{\log m}{m}} \right), \mathcal{Z}_{a,b} > f\left(T_0\right) \right) \\
\underset{\mathrm{(\ddag)}}{\leqslant}\ &\sum_{m=T_0}^{l} \sum_{a,b\in\mathcal{A}, a<b} \mathbb{P}\left( \sum_{j=1}^{m}\left( X_{a,j} - X_{b,j} -\delta \right) < -4\sqrt{m\log m}, \mathcal{Z}_{a,b} > f\left(T_0\right) \right) \\
=\ &\bar{\Phi}\left( f\left(T_0\right) \right)\sum_{m=T_0}^{l} \sum_{a,b\in\mathcal{A}, a<b} \mathbb{P}\left( \sum_{j=1}^{m}\left( X_{a,j} - X_{b,j} -\delta \right) < -4\sqrt{m\log m}\right) \\
\underset{\mathrm{(\star)}}{\leqslant}\ &\bar{\Phi}\left( f\left(T_0\right) \right)\sum_{m=T_0}^{l}\sum_{a,b\in\mathcal{A}, a<b}\frac{1}{m^8} \\
\leqslant\ &\bar{\Phi}\left( f\left(T_0\right) \right)K^2\sum_{m=T_0}^{\infty}\frac{1}{m^8} \\
\underset{\mathrm{(\ast)}}{\leqslant}\ &\frac{\bar{\Phi}\left( f\left(T_0\right) \right)}{2},
\end{align*}
where $(\dag)$ follows from Lemma~\ref{lemma:prop1K}, $(\ddag)$ from Lemma~\ref{lemma:prop2K}, $(\star)$ follows using the Chernoff-Hoeffding bound \\citep{hoeffding} and finally, $(\ast)$ follows from the definition of $T_0$.
Therefore, we have
\begin{align*}
&\mathbb{P}\left(\tau \leqslant l \right) \leqslant \frac{\bar{\Phi}\left( f\left(T_0\right) \right)}{2} + \Phi\left( f\left(T_0\right) \right) = 1 - \frac{\bar{\Phi}\left( f\left(T_0\right) \right)}{2} \\
\implies\ & \mathbb{P}\left(\tau > l \right) \geqslant \frac{\bar{\Phi}\left( f\left(T_0\right) \right)}{2}.
\end{align*}
Taking the limit $l\to\infty$ and appealing to the continuity of probability, we obtain
\begin{align*}
&\mathbb{P}\left(\tau =\infty \right) \geqslant \frac{\bar{\Phi}\left( f\left(T_0\right) \right)}{2} \\
\implies\ &\mathbb{P}\left( \bigcap_{m \geqslant 1}\bigcap_{a,b\in\mathcal{A}, a<b}\left\lbrace \left\lvert \mathcal{Z}_{a,b} + \sum_{j=1}^{m}\left( X_{a,j} - X_{b,j} \right) \right\rvert \geqslant 4\sqrt{m\log m} \right\rbrace \right) \geqslant \frac{\bar{\Phi}\left( f\left(T_0\right) \right)}{2}.
\end{align*} \hfill $\square$

\section{Proof of Theorem~\ref{thm:alg2}}
\label{proof:alg2}

We will initially assume $\delta > 8\sqrt{\log n/n}$ for technical convenience. In the final step leading up to the asserted bound, we will relax this assumption by offsetting regret appropriately.

Let $\mathcal{A} := \{1,...,K\}$. Define the following stopping times:
\begin{align*}
\tau_1\left( s_n \right) &:= \inf \left\lbrace m \geqslant s_n : \exists a,b\in\mathcal{A}, a<b\ \text{s.t.}\ \left\lvert \mathcal{Z}_{a,b} + \sum_{j=1}^{m} \left( X_{a,j} - X_{b,j} \right) \right\rvert < 4\sqrt{m\log m} \right\rbrace, \\
\tau_2\left( s_n \right) &:= \inf \left\lbrace m \geqslant s_n : \left\lvert \sum_{j=1}^{m} \left( X_{a,j} - X_{b,j} \right) \right\rvert \geqslant 4\sqrt{m\log n}\ \forall\ a,b\in\mathcal{A},\ a<b \right\rbrace.
\end{align*}

To keep notations simple, we will suppress the argument and denote $\tau_1\left( s_n \right)$ and $\tau_2\left( s_n \right)$ by $\tau_1$ and $\tau_2$ respectively (the dependence on $s_n$ will be implicit going forward). Let $R_t$ denote the cumulative pseudo-regret of \texttt{ALG2}$(n)$ after $t \leqslant n$ pulls. Let \texttt{D} denote the event that the first batch of $K$ arms queried from the reservoir is ``all-distinct,'' i.e., no two arms in this batch belong to the same type; let \texttt{D\textsuperscript{c}} be the complement of this event. Let \texttt{CI} denote the event that the algorithm commits to an inferior-typed arm. Let $\tilde{R}_{\cdot},\bar{R}_{\cdot}$ be independently drawn from the same distribution as $R_{\cdot}$. Let $x^+ := \max(x,0)$ for $x\in\mathbb{R}$. Then, $R_n$ evolves according to the following stochastic recursion:
\begin{align*}
&R_n \\
\leqslant\ &\mathbbm{1}\{\texttt{D}\}\left[ \mathbbm{1}\left\lbrace \tau_1 < \tau_2 \right\rbrace \left( \bar{\Delta}\min\left(K\tau_1, n\right) + \tilde{R}_{\left( n-K\tau_1 \right)^+} \right) + \mathbbm{1}\left\lbrace \tau_1 \geqslant \tau_2 \right\rbrace \left( \bar{\Delta}\min\left(K\tau_2, n\right) + \mathbbm{1}\{\texttt{CI}\}\bar{\Delta}\left( n - K\tau_2 \right)^+ \right) \right] \\
&+ \mathbbm{1}\{\texttt{D\textsuperscript{c}}\} \left[ \mathbbm{1}\left\lbrace \tau_1 < \tau_2 \right\rbrace \left( \bar{\Delta}\min\left(K\tau_1, n\right) + \bar{R}_{\left( n-K\tau_1 \right)^+} \right) + \mathbbm{1}\left\lbrace \tau_1 \geqslant \tau_2 \right\rbrace \bar{\Delta} n \right] \\
\leqslant\ &\mathbbm{1}\{\texttt{D}\}\left[ \mathbbm{1}\left\lbrace \tau_1 < \tau_2 \right\rbrace \left( \bar{\Delta}\min\left(K\tau_2, n\right) + \tilde{R}_n \right) + \mathbbm{1}\left\lbrace \tau_1 \geqslant \tau_2 \right\rbrace \left( \bar{\Delta}\min\left(K\tau_2, n\right) + \mathbbm{1}\{\texttt{CI}\}\bar{\Delta} n \right) \right] \\
&+ \mathbbm{1}\{\texttt{D\textsuperscript{c}}\} \left[ \mathbbm{1}\left\lbrace \tau_1 < \tau_2 \right\rbrace \left( \bar{\Delta}\min\left(K\tau_1, n\right) + \bar{R}_n \right) + \mathbbm{1}\left\lbrace \tau_1 \geqslant \tau_2 \right\rbrace \bar{\Delta} n \right] \\
\leqslant\ &\mathbbm{1}\{\texttt{D}\}\left[ 2\bar{\Delta}\min\left(K\tau_2, n\right) + \mathbbm{1}\left\lbrace \tau_1 < \tau_2 \right\rbrace\tilde{R}_n + \mathbbm{1}\{\texttt{CI}\}\bar{\Delta} n \right] + \mathbbm{1}\{\texttt{D\textsuperscript{c}}\} \left[ \bar{\Delta}K\tau_1 + \bar{R}_n + \mathbbm{1}\left\lbrace \tau_1 \geqslant \tau_2 \right\rbrace \bar{\Delta} n \right].
\end{align*}

Taking expectations on both sides, one recovers using the independence of $\tilde{R}_n,\bar{R}_n$ that
\begin{align*}
&\mathbb{E}R_n \\
\leqslant\ &\frac{\bar{\Delta}}{\mathbb{P}\left( \left. \tau_1 \geqslant \tau_2 \right\rvert \texttt{D} \right)} \left[ 2\mathbb{E}\left[ \left. \min\left(K\tau_2,n\right) \right\rvert \texttt{D} \right] + \left(\frac{\mathbb{P}\left( \texttt{D\textsuperscript{c}} \right)}{\mathbb{P}\left( \texttt{D} \right)}\right)\mathbb{E}\left[ \left. K\tau_1 \right\rvert \texttt{D\textsuperscript{c}} \right] + \left( \mathbb{P}\left( \left. \texttt{CI} \right\rvert \texttt{D} \right) + \left(\frac{\mathbb{P}\left( \texttt{D\textsuperscript{c}} \right)}{\mathbb{P}\left( \texttt{D} \right)}\right)\mathbb{P}\left( \left. \tau_1 \geqslant \tau_2 \right\rvert \texttt{D\textsuperscript{c}} \right) \right) n \right],
\end{align*}
where $\mathbb{P}\left( \texttt{D} \right) = K!\prod_{i=1}^K\alpha_i$.

\subsection{Lower bounding $\mathbb{P}\left( \left. \tau_1 \geqslant \tau_2 \right\rvert \mathtt{D} \right)$}

Define the following:
\begin{align}
\gamma\left(s_n\right) := \mathbb{P}\left( \left. \tau_1\left( s_n \right) = \infty\right\rvert \mathtt{D} \right), \label{eqn:gamma-scaling}
\end{align}
where $\tau_1\left( s_n \right)$ and $\mathtt{D}$ are as defined before. We will suppress the dependence on $s_n$ to keep notations minimal. Note that 
\begin{align*}
\mathbb{P}\left( \left. \tau_1 < \tau_2 \right\rvert \mathtt{D} \right) &= \mathbb{P}\left( \left. \tau_1 < \tau_2, \tau_2 = \infty\right\rvert \mathtt{D} \right) + \mathbb{P}\left( \left. \tau_1 < \tau_2, \tau_2 < \infty \right\rvert \mathtt{D} \right) \\
&\leqslant \mathbb{P}\left( \left. \tau_2 = \infty\right\rvert \mathtt{D} \right) + \mathbb{P}\left( \left. \tau_1 < \infty\right\rvert \mathtt{D} \right) \\
&= \mathbb{P}\left( \left. \tau_1 < \infty\right\rvert \mathtt{D} \right) \\
&= 1 - \gamma\left(s_n\right),
\end{align*}
where the equality in the third step follows since $\tau_2$ is almost surely finite on the event \texttt{D} (proved in \S\ref{subsubsec:rando} below), and the final equality is due to \eqref{eqn:gamma-scaling}. Thus, $\mathbb{P}\left( \left. \tau_1 \geqslant \tau_2 \right\rvert \mathtt{D} \right) \geqslant \gamma\left(s_n\right)$.

\subsubsection{Proof that $\tau_2$ is almost surely finite on \texttt{D}}
\label{subsubsec:rando}

Let $\mathbb{P}_{\texttt{D}}(\cdot) := \mathbb{P}\left( \cdot | \texttt{D} \right)$ be the conditional measure w.r.t. the event \texttt{D}. Let $\mathcal{A} := \{1,...,K\}$. Then, by continuity of probability, we have
\begin{align*}
\mathbb{P}_\texttt{D}\left( \tau_2 = \infty \right) &= \lim_{l\to\infty}\mathbb{P}_\texttt{D}\left( \tau_2 > l \right) \\
&= \lim_{l\to\infty}\mathbb{P}_\texttt{D}\left( \bigcap_{m=s_n}^{l}\bigcup_{a,b\in\mathcal{A}, a<b} \left\lbrace \left\lvert \sum_{j=1}^{m} \left( X_{a,j} - X_{b,j} \right) \right\rvert \ < 4\sqrt{m\log n} \right\rbrace \right) \\
&\leqslant \lim_{l\to\infty} \sum_{a,b\in\mathcal{A}, a<b}\mathbb{P}_\texttt{D}\left( \left\lvert \sum_{j=1}^{l} \left( X_{a,j} - X_{b,j} \right) \right\rvert \ < 4\sqrt{l\log n} \right).
\end{align*}

On \texttt{D}, it must be that $\left\lvert \mathbb{E}\left[ X_{a,j} - X_{b,j} \right] \right\rvert \geqslant \delta$. Without loss of generality, assume that $\mathbb{E}\left[ X_{a,j} - X_{b,j} \right] \geqslant \delta$. Then,

\begin{align*}
\mathbb{P}_\texttt{D}\left( \tau_2 = \infty \right) &\leqslant \lim_{l\to\infty}\sum_{a,b\in\mathcal{A}, a<b}\mathbb{P}_\texttt{D}\left( \sum_{j=1}^{l} \left( X_{a,j} - X_{b,j} \right) < 4\sqrt{l\log n} \right) \\
&= \lim_{l\to\infty}\sum_{a,b\in\mathcal{A}, a<b}\mathbb{P}_\texttt{D}\left( \sum_{j=1}^{l} \left( X_{a,j} - X_{b,j} -\delta \right) < -l\left( \delta - 4\sqrt{\frac{\log n}{l}} \right) \right) \\
&\leqslant \lim_{l\to\infty}\sum_{a,b\in\mathcal{A}, a<b}\mathbb{P}_\texttt{D}\left( \sum_{j=1}^{l} \left( X_{a,j} - X_{b,j} - \delta \right) < -4l\sqrt{\log n}\left( \frac{2}{\sqrt{n}} - \frac{1}{\sqrt{l}} \right) \right),
\end{align*}

where the last inequality follows since $\delta > 8\sqrt{\log n/n}$ (by assumption). Now, using the Chernoff-Hoeffding bound \citep{hoeffding} together with the fact that $-1 \leqslant X_{a,j} - X_{b,j} \leqslant 1$, we obtain for $l > n$ and any $a,b\in\mathcal{A}, a<b$ that 

\begin{align*}
\mathbb{P}_\texttt{D}\left( \sum_{j=1}^{l} \left( X_{a,j} - X_{b,j} -\delta \right) < -4l\sqrt{\log n}\left( \frac{2}{\sqrt{n}} - \frac{1}{\sqrt{l}} \right) \right) &\leqslant \exp\left[ -8l\left( \frac{2}{\sqrt{n}} - \frac{1}{\sqrt{l}} \right)^2 \log n \right] \\
&= \exp\left[ -8\left( \frac{4l}{n} - 4\sqrt{\frac{l}{n}} +1 \right)^2 \log n \right].
\end{align*}
Summing over $a,b\in\mathcal{A}, a<b$ and taking the limit $l\to\infty$ proves the stated assertion. \hfill $\square$

\subsection{Upper bounding $\mathbb{E}\left[ \left. \min\left(K\tau_2,n\right) \right\rvert \mathtt{D} \right]$}

Let $\mathbb{P}_{\texttt{D}}(\cdot) := \mathbb{P}\left( \cdot | \texttt{D} \right)$ be the conditional measure w.r.t. the event \texttt{D}. Let $\mathcal{A} := \{1,...,K\}$. Then,

\begin{align*}
\mathbb{E}\left[ \left. \min\left(K\tau_2,n\right) \right\rvert \texttt{D} \right] &= K\mathbb{E}\left[ \left. \min\left(\tau_2,\frac{n}{K}\right) \right\rvert \texttt{D} \right] \\
&\leqslant K\mathbb{E}\left[ \left. \min\left(\tau_2,n\right) \right\rvert \texttt{D} \right] \\
&\leqslant Ks_n + K\sum_{k=s_n+1}^{n} \mathbb{P}_\texttt{D}\left( \tau_2 \geqslant k \right) \\
&\leqslant Ks_n + K\sum_{k=s_n}^{n}\mathbb{P}_\texttt{D}\left( \tau_2 \geqslant k+1 \right) \\
&\leqslant Ks_n + K\sum_{k=1}^{n} \sum_{a,b\in\mathcal{A}, a<b}\mathbb{P}_\texttt{D}\left( \left\lvert \sum_{j=1}^{k}\left( X_{a,j} - X_{b,j} \right) \right\rvert < 4\sqrt{k\log n} \right).
\end{align*}

On \texttt{D}, it must be that $\left\lvert \mathbb{E}\left[ X_{a,j} - X_{b,j} \right] \right\rvert \geqslant \delta$. Without loss of generality, assume that $\mathbb{E}\left[ X_{a,j} - X_{b,j} \right] \geqslant \delta$. Then,
\begin{align*}
\mathbb{E}\left[ \left. \min\left(K\tau_2,n\right) \right\rvert \texttt{D} \right] &\leqslant Ks_n + K\sum_{k=1}^{n} \sum_{a,b\in\mathcal{A}, a<b} \mathbb{P}_\texttt{D}\left( \sum_{j=1}^{k} \left( X_{a,j} - X_{b,j} \right) < 4\sqrt{k\log n} \right) \\
&= Ks_n + K\sum_{k=1}^{n} \sum_{a,b\in\mathcal{A}, a<b} \mathbb{P}_\texttt{D}\left( \sum_{j=1}^{k} \left( X_{a,j} - X_{b,j} - \delta \right) < -k\left( \delta - 4\sqrt{\frac{\log n}{k}} \right) \right) \\
&\leqslant Ks_n + \frac{32K^3\log n}{{\delta}^2} + K\sum_{k = \left\lceil \frac{64\log n}{{\delta}^2} \right\rceil}^{n} \sum_{a,b\in\mathcal{A}, a<b} \mathbb{P}_\texttt{D}\left( \sum_{j=1}^{k} \left( X_{a,j} - X_{b,j} -\delta \right) < \frac{-k\delta}{2} \right),
\end{align*}
where the last step follows since $\delta > 8\sqrt{\log n/n}$ (by assumption) implies $n > 64\log n/{\delta}^2$, and $k \geqslant 64\log n/{\delta}^2$ implies $\delta  - 4\sqrt{{\log n}/{k}} \geqslant \delta/2$. Finally, using the Chernoff-Hoeffding inequality \citep{hoeffding} together with the fact that $\left\lvert X_{a,j} - X_{b,j} \right\rvert \leqslant 1$, one obtains
\begin{align*}
\mathbb{E}\left[ \left. \min\left(K\tau_2,n\right) \right\rvert \texttt{D} \right] \leqslant Ks_n + \frac{32K^3\log n}{{\delta}^2} + \frac{K^3}{2}\sum_{k = \left\lceil \frac{64\log n}{{\delta}^2} \right\rceil}^{n} \exp\left( \frac{-{{\delta}}^2k}{8} \right) \leqslant Ks_n + \frac{64K^3\log n}{{\delta}^2}.
\end{align*}

\subsection{Upper bounding $\mathbb{E}\left[ \left. K\tau_1 \right\rvert \mathtt{D\textsuperscript{c}} \right]$}

The event \texttt{D\textsuperscript{c}} will be implicitly assumed and we will drop the conditional argument for notational simplicity. Let $\mathcal{A} := \{1,...,K\}$. Without loss of generality, suppose that arm~$1$ and $2$ belong to the same type. Then,

\begin{align*}
\mathbb{E}\left[ \left. K\tau_1 \right\rvert \mathtt{D\textsuperscript{c}} \right] &= Ks_n + K\sum_{k \geqslant s_n+1} \mathbb{P}\left( \tau_1 \geqslant k \right) \\
&= Ks_n + K\sum_{k \geqslant s_n} \mathbb{P}\left( \tau_1 \geqslant k+1 \right) \\
&= Ks_n+K\sum_{k \geqslant s_n} \mathbb{P}\left( \bigcap_{m=s_n}^{k} \bigcap_{a,b\in\mathcal{A}, a<b}\left\lbrace \left\lvert \mathcal{Z}_{a,b} + \sum_{j=1}^{m}\left( X_{1,j} - X_{2,j} \right) \right\rvert \geqslant 4\sqrt{m\log m} \right\rbrace \right) \\
&\leqslant Ks_n+K\sum_{k \geqslant 1} \mathbb{P}\left( \left\lvert \mathcal{Z}_{1,2} + \sum_{j=1}^{k}\left( X_{1,j} - X_{2,j} \right) \right\rvert \geqslant 4\sqrt{k\log k} \right).
\end{align*}

Since $\mathcal{Z}_{a,b}$ is a standard Gaussian, and the increments $X_{1,j} - X_{2,j}$ are zero-mean sub-Gaussian with variance proxy $1$, it follows from the Chernoff-Hoeffding concentration bound \citep{hoeffding} that
\begin{align*}
\mathbb{E}\left[ \left. K\tau_1 \right\rvert \mathtt{D\textsuperscript{c}} \right] &\leqslant Ks_n + 2K\sum_{k \geqslant 1}\frac{1}{k^4} = \left(s_n + \frac{\pi^4}{45}\right)K < \left(s_n+3\right)K.
\end{align*} 

\subsection{Upper bounding $\mathbb{P}\left( \left. \mathtt{CI}\right\rvert \mathtt{D} \right)$}

Let $\mathbb{P}_{\texttt{D}}(\cdot) := \mathbb{P}\left( \cdot | \texttt{D} \right)$ be the conditional measure w.r.t. the event \texttt{D}. Let $\mathcal{A} := \{1,...,K\}$ and without loss of generality, suppose that arm~$1$ is optimal (mean $\mu_1$). Then,

\begin{align*}
\mathbb{P}\left( \left. \mathtt{CI}\right\rvert \mathtt{D} \right) &\leqslant \mathbb{P}_\texttt{D} \left( \bigcup_{b=2}^{K} \left\lbrace \sum_{j=1}^{\tau_2}\left( X_{1,j} - X_{b,j} \right) \leqslant -4\sqrt{\tau_2\log n} \right\rbrace \right) \\
&\leqslant \sum_{b=2}^{K} \sum_{k=s_n}^{n} \mathbb{P}_\texttt{D} \left( \sum_{j=1}^{k}\left( X_{1,j} - X_{b,j} \right) \leqslant -4\sqrt{k\log n} \right) + \sum_{b=2}^{K}\mathbb{P}_\texttt{D}\left( \tau_2 > n \right) \\
&\leqslant \sum_{b=2}^{K} \sum_{k=1}^{n} \frac{1}{n^{8}} + \frac{K^3}{n^8} \\
&\leqslant \frac{K}{n^7} + \frac{K^3}{n^8},
\end{align*}
where the second-to-last step follows using the Chernoff-Hoeffding inequality \citep{hoeffding}.

\subsection{Upper bounding $\mathbb{P}\left( \left. \tau_1 \geqslant \tau_2 \right\rvert \mathtt{D\textsuperscript{c}} \right)$}

Let $\mathbb{P}_{\texttt{D\textsuperscript{c}}}(\cdot) := \mathbb{P}\left( \cdot | \texttt{D\textsuperscript{c}} \right)$ be the conditional measure w.r.t. the event \texttt{D\textsuperscript{c}}. Let $\mathcal{A} := \{1,...,K\}$. On \texttt{D\textsuperscript{c}}, there exist $2$ arms in $\mathcal{A}$ that belong to the same type; without loss of generality suppose that these arms are indexed by $1,2$. Then,

\begin{align}
\mathbb{P}\left( \left. \tau_1 \geqslant \tau_2 \right\rvert \mathtt{D\textsuperscript{c}} \right) &\leqslant \mathbb{P}\left( \left. \tau_1 > n \right\rvert \mathtt{D\textsuperscript{c}} \right) + \mathbb{P}\left( \left. \tau_2 \leqslant n \right\rvert \mathtt{D\textsuperscript{c}} \right) \notag \\
&\leqslant \frac{2}{n^4} + \mathbb{P}_{\texttt{D\textsuperscript{c}}}\left( \tau_2 \leqslant n \right) \notag \\
&= \frac{2}{n^4} + \mathbb{P}_{\texttt{D\textsuperscript{c}}} \left( \bigcup_{m=s_n}^{n} \bigcap_{a,b\in\mathcal{A}, a<b} \left\lbrace \left\lvert \sum_{j=1}^{m} \left( X_{a,j} - X_{b,j} \right) \right\rvert \geqslant 4\sqrt{m\log n} \right\rbrace \right) \notag \\
&\leqslant \frac{2}{n^4} + \sum_{m=1}^{n}\mathbb{P}_{\texttt{D\textsuperscript{c}}} \left( \left\lvert \sum_{j=1}^{m} \left( X_{1,j} - X_{2,j} \right) \right\rvert \geqslant 4\sqrt{m\log n} \right) \notag \\
&\leqslant \frac{2}{n^4} + \frac{2}{n^7},
\end{align}
where the last step follows using the Chernoff-Hoeffding bound \citep{hoeffding}.

\subsection{Putting everything together}

Combining everything, one finally obtains that when $\delta > 8\sqrt{\log n/ n}$,
\begin{align*}
\mathbb{E}R_n \leqslant \frac{CK^3\bar{\Delta}}{\gamma\left(s_n\right)}\left( \frac{\log n}{{\delta}^2} + \frac{s_n}{\mathbb{P}\left(\texttt{D}\right)} \right),
\end{align*} 
where $\gamma\left(s_n\right)$ is as defined in \eqref{eqn:gamma-scaling}, $\mathbb{P}\left( \texttt{D} \right) = K!\prod_{i=1}^K\alpha_i$, and $C$ is some absolute constant. When $\delta \leqslant 8\sqrt{\log n/ n}$, regret is at most $\bar{\Delta}n \leqslant 64\bar{\Delta}/\delta^2 \log n$. Thus, the aforementioned bound, in fact, holds generally for some large enough absolute constant $C$. \hfill $\square$

\section{Auxiliary results used in the analysis of \texttt{\textbf{ALG3}}}
\label{appendix:aux_alg3}

\begin{lemma}[Persistence of heterogeneous consideration sets]
\label{lemma:alg3_1}
Consider a two-armed bandit with rewards bounded in $[0,1]$, means $\mu_1 > \mu_2$, and gap $\underline{\Delta}=\mu_1 - \mu_2$. Let $\left\lbrace X_{i,j} : j=1,2,... \right\rbrace$ denote the sequence of rewards collected from arm~$i\in\{1,2\}$ by \texttt{UCB1} \citep{auer2002}. Let $\mathcal{Z}$ be an independently generated standard Gaussian random variable. Let $\left( N_1(n), N_2(n) \right)$ be the per-arm sample counts under \texttt{UCB1} up to and including time $n$. Define
\begin{align*}
&M_n := \min\left( N_1(n), N_2(n) \right), \notag \\
&\tau := \inf \left\lbrace n\in\mathbb{N} : \left\lvert \mathcal{Z} + \sum_{j=1}^{M_n} \left( X_{1,j} - X_{2,j} \right) \right\rvert < 4\sqrt{M_n\log M_n} \right\rbrace.
\end{align*}
Then, $\mathbb{P}\left( \tau = \infty \right) > \beta_{\underline{\Delta},2}$, where $\beta_{\underline{\Delta},2}$ is as defined in \eqref{eqn:beta_K} with $\delta\gets\underline{\Delta}$ and $K\gets 2$.
\end{lemma}

\begin{lemma}[Fast rejection of homogeneous consideration sets]
\label{lemma:alg3_2}
Consider a two-armed bandit where both arms have equal means. Let $\left\lbrace X_{i,j} : j=1,2,... \right\rbrace$ denote the sequence of rewards collected from arm~$i\in\{1,2\}$ by \texttt{UCB1} \citep{auer2002}. Let $\mathcal{Z}$ be an independently generated standard Gaussian random variable. Let $\left( N_1(n), N_2(n) \right)$ be the per-arm sample counts under \texttt{UCB1} up to and including time $n$. Define
\begin{align*}
&M_n := \min\left( N_1(n), N_2(n) \right), \notag \\
&\tau := \inf \left\lbrace n\in\mathbb{N} : \left\lvert \mathcal{Z} + \sum_{j=1}^{M_n} \left( X_{1,j} - X_{2,j} \right) \right\rvert < 4\sqrt{M_n\log M_n} \right\rbrace.
\end{align*}
Then, there exists an absolute constant $C$ such that $\mathbb{E}\tau \leqslant C$.
\end{lemma}

\subsection{Proof of Lemma~\ref{lemma:alg3_1}}

Since the rewards are uniformly bounded in $[0,1]$, it follows that $N_i(n) \to \infty$ for each arm~$i\in\{1,2\}$ as $n\to\infty$ on every sample-path. This is due to the structure of the upper confidence bounds used by \texttt{UCB1}. Consequently, $M_n = \min \left( N_1(n), N_2(n) \right) \to \infty$ as $n\to\infty$ on every sample-path. Also note that $M_n$ is a weakly increasing integer-valued process (starting from $1$) with unit increments, wherever they exist. Thus, it follows on every sample-path that $\tau$, in fact, weakly dominates the stopping time $\tau^\prime$ defined below
\begin{align}
\tau^\prime := \inf \left\lbrace m\in\mathbb{N} : \left\lvert \mathcal{Z} + \sum_{j=1}^{m} \left( X_{1,j} - X_{2,j} \right) \right\rvert < 4\sqrt{m\log m} \right\rbrace.
\end{align}
Therefore, $\mathbb{P}\left( \tau = \infty \right) \geqslant \mathbb{P}\left( \tau^\prime = \infty \right) > \beta_{\underline{\Delta},2}$, where the last inequality follows from Proposition~\ref{propK} with $\delta\gets\underline{\Delta}$ and $K\gets 2$. \hfill $\square$

\newpage

\subsection{Proof of Lemma~\ref{lemma:alg3_2}}

Note that
\begin{align*}
\mathbb{E}\tau &= 1 + \sum_{k \geqslant 2}\mathbb{P}\left( \tau \geqslant k \right) \\
&= 1 + \sum_{k \geqslant 2}\mathbb{P}\left( \bigcap_{n=1}^{k-1} \left\lbrace  \left\lvert \mathcal{Z} + \sum_{j=1}^{M_n} \left( X_{1,j} - X_{2,j} \right) \right\rvert \geqslant 4\sqrt{M_n\log M_n}  \right\rbrace \right) \\
&\leqslant 1 + \sum_{k \geqslant 1}\mathbb{P}\left( \left\lvert \mathcal{Z} + \sum_{j=1}^{M_k} \left( X_{1,j} - X_{2,j} \right) \right\rvert \geqslant 4\sqrt{M_k\log M_k} \right) \\
&= 1 + \sum_{k \geqslant 1}\sum_{m = 1}^{k}\mathbb{P}\left( \left\lvert \mathcal{Z} + \sum_{j=1}^{M_k} \left( X_{1,j} - X_{2,j} \right) \right\rvert \geqslant 4\sqrt{M_k\log M_k}, N_1(k) = m \right) \\
&= 1 + \sum_{k \geqslant 1}\sum_{1 \leqslant m \leqslant k/2}\mathbb{P}\left( \left\lvert \mathcal{Z} + \sum_{j=1}^{m} \left( X_{1,j} - X_{2,j} \right) \right\rvert \geqslant 4\sqrt{m\log m}, N_1(k) = m \right) \\
&{\color{white}=} + \sum_{k \geqslant 1}\sum_{k/2 < m \leqslant k}\mathbb{P}\left( \left\lvert \mathcal{Z} + \sum_{j=1}^{(k-m)} \left( X_{1,j} - X_{2,j} \right) \right\rvert \geqslant 4\sqrt{(k-m)\log (k-m)}, N_1(k) = m \right) \\ 
&= 1 + \sum_{k \geqslant 1}\sum_{1 \leqslant m \leqslant k/2}\mathbb{P}\left( \left\lvert \mathcal{Z} + \sum_{j=1}^{m} \left( X_{1,j} - X_{2,j} \right) \right\rvert \geqslant 4\sqrt{m\log m}, N_1(k) = m \right) \\
&{\color{white}=} + \sum_{k \geqslant 1}\sum_{1 \leqslant m < k/2}\mathbb{P}\left( \left\lvert \mathcal{Z} + \sum_{j=1}^{m} \left( X_{1,j} - X_{2,j} \right) \right\rvert \geqslant 4\sqrt{m\log m}, N_2(k) = m \right) \\ 
&\leqslant 1 + 2\sum_{k \geqslant 1}\sum_{\theta k \leqslant m \leqslant k/2}\mathbb{P}\left( \left\lvert \mathcal{Z} + \sum_{j=1}^{m} \left( X_{1,j} - X_{2,j} \right) \right\rvert \geqslant 4\sqrt{m\log m} \right) \\
&{\color{white}=} + \sum_{k \geqslant 1}\left[ \mathbb{P}\left( N_1(k) \leqslant \theta k \right) + \mathbb{P}\left( N_2(k) \leqslant \theta k \right) \right],
\end{align*}
where $\theta = 1/2 - \sqrt{15}/8$. Using Theorem~4(i) of \cite{kalvit2020finite} with $\epsilon = \sqrt{15}/8$, one obtains
\begin{align*}
\mathbb{E}\tau &\leqslant 1 + 2\sum_{k \geqslant 1}\sum_{\theta k \leqslant m \leqslant k/2}\mathbb{P}\left( \left\lvert \mathcal{Z} + \sum_{j=1}^{m} \left( X_{1,j} - X_{2,j} \right) \right\rvert \geqslant 4\sqrt{m\log m} \right) + 16\sum_{k \geqslant 1}\frac{1}{k^2} \\
&\leqslant 1 + 4\sum_{k \geqslant 1}\sum_{\theta k \leqslant m \leqslant k/2}\frac{1}{m^4} + 16\sum_{k \geqslant 1}\frac{1}{k^2} \\
&\leqslant 1 + \frac{4}{\theta^4}\sum_{k \geqslant 1}\frac{1}{k^3} + 16\sum_{k \geqslant 1}\frac{1}{k^2}.
\end{align*} \hfill $\square$

\section{Proof of Theorem~\ref{thm:alg3}}
\label{proof:alg3}

Consider the first epoch and define the following:
\begin{align}
&M_n := \min\left( N_1(n), N_2(n) \right), \notag \\
&\tau := \inf \left\lbrace n\in\mathbb{N} : \left\lvert \mathcal{Z} + \sum_{j=1}^{M_n} \left( X_{1,j} - X_{2,j} \right) \right\rvert < 4\sqrt{M_n\log M_n} \right\rbrace, \notag
\end{align}
where $\left( N_1(n), N_2(n) \right)$ are the per-arm sample counts under \texttt{UCB1} up to and including time $n$. Note that $\tau$ marks the termination of epoch~$1$. 

Let $R_n$ denote the cumulative pseudo-regret of \texttt{ALG3} after $n$ pulls (superscript $\pi$ suppressed for notational convenience). Let $S_n$ denote the cumulative pseudo-regret of \texttt{UCB1} after $n$ pulls in a two-armed bandit with gap $\underline{\Delta}$. Let \texttt{D} and \texttt{I} respectively denote the events that the two arms queried in epoch~$1$ have distinct and identical types. Similarly, let \texttt{OPT} and \texttt{INF} respectively denote the events that the two arms have ``optimal'' and ``inferior'' types (Note that $\texttt{I} = \texttt{OPT}\cup\texttt{INF}$). Let $\tilde{R}_n,\bar{R}_n,\hat{R}_n$ be independently drawn from the same distribution as $R_n$. Then, note that $R_n$ admits the following stochastic evolution:

\begin{align*}
R_n &= \mathbbm{1}\{\texttt{D}\}\left[ S_{\min\left(\tau, n\right)} + \tilde{R}_{\left( n-\tau \right)^+} \right] + \mathbbm{1}\{\texttt{INF}\} \left[ \underline{\Delta}\min\left(\tau, n\right) + \bar{R}_{\left( n-\tau \right)^+} \right] + \mathbbm{1}\{\texttt{OPT}\}\hat{R}_{\left( n-\tau \right)^+} \\
&\leqslant \mathbbm{1}\{\texttt{D}\}\left[ S_{n} + \mathbbm{1}\left\lbrace \tau < n \right\rbrace\tilde{R}_{n} \right] + \mathbbm{1}\{\texttt{INF}\} \left[ \underline{\Delta}\tau + \bar{R}_{n} \right] + \mathbbm{1}\{\texttt{OPT}\}\hat{R}_{n} \\
&\leqslant \mathbbm{1}\{\texttt{D}\}\left[ S_{n} + \mathbbm{1}\left\lbrace \tau < \infty \right\rbrace\tilde{R}_{n} \right] + \mathbbm{1}\{\texttt{INF}\} \left[ \underline{\Delta}\tau + \bar{R}_{n} \right] + \mathbbm{1}\{\texttt{OPT}\}\hat{R}_{n},
\end{align*}
where the first inequality follows since \texttt{ALG3} is agnostic to $n$, and hence the pseudo-regret $R_n$ is weakly increasing in $n$. Taking expectations on both sides, one recovers using the independence of $\tilde{R}_n,\bar{R}_n,\hat{R}_n$ that

\begin{align*}
\mathbb{E}R_n &\leqslant \frac{1}{\mathbb{P}\left( \left. \tau = \infty \right\rvert \texttt{D} \right)}\left[ \mathbb{E}S_n + \left( \frac{1-\alpha_1}{2\alpha_1} \right) \underline{\Delta} \mathbb{E}\left[ \left.\tau \right\rvert \texttt{INF} \right] \right] \\
&\leqslant \frac{1}{\beta_{\underline{\Delta},2}}\left[ \mathbb{E}S_n + C\left( \frac{1-\alpha_1}{2\alpha_1} \right) \underline{\Delta} \right],
\end{align*}
where $\beta_{\underline{\Delta},2}$ is as defined in \eqref{eqn:beta_K} with $\delta\gets\underline{\Delta}$ and $K\gets 2$, and $C$ is some absolute constant; the last inequality follows using Lemma~\ref{lemma:alg3_1} and \ref{lemma:alg3_2}. The stated assertion now follows since $\mathbb{E}S_n \leqslant C^\prime\left( \log n/\underline{\Delta} + \underline{\Delta} \right)$ for some absolute constant $C^\prime$ \citep{auer2002}. \hfill $\square$

\section{Auxiliary results used in the analysis of \texttt{\textbf{ALG4}}}
\label{appendix:aux_alg4}

\begin{lemma}[Persistence of heterogeneous consideration sets]
\label{lemma:alg4_1}
Consider a $K$-armed bandit with rewards bounded in $[0,1]$ and means $\mu_1 > ... > \mu_K$. Let $\left\lbrace X_{a,j} : j=1,2,... \right\rbrace$ denote the rewards collected from arm~$a\in\{1,...,K\} =: \mathcal{A}$ by \texttt{UCB1} \citep{auer2002}. Let $\left\lbrace \mathcal{Z}_{a,b} : a,b\in\mathcal{A}, a<b \right\rbrace$ be a collection of ${K \choose 2}$ independent standard Gaussian random variables. Let $N_a(n)$ be the sample count of arm~$a$ under \texttt{UCB1} until time $n$. Define
\begin{align}
&M_l := \min_{a\in\mathcal{A}} N_a(l), \notag \\
&\tau := \inf \left\lbrace l \geqslant K : \exists\ a,b\in\mathcal{A}, a<b\ \text{s.t.}\ \left\lvert \mathcal{Z}_{a,b} + \sum_{j=1}^{M_l} \left( X_{a,j} - X_{b,j} \right) \right\rvert < 4\sqrt{M_l\log M_l} \right\rbrace. \notag
\end{align}
Then, $\mathbb{P}\left( \tau = \infty \right) > \beta_{\delta,K}$, where $\beta_{\delta,K}$ is as defined in \eqref{eqn:beta_K}.
\end{lemma}

\begin{lemma}[Path-wise lower bound on the arm-sampling rate of \texttt{\textbf{UCB1}}]
\label{lemma:alg4_2}
Consider a $K$-armed bandit with rewards bounded in $[0,1]$. Let $N_a(n)$ be the sample count of arm~$a\in\{1,...,K\} =: \mathcal{A}$ under \texttt{UCB1} \citep{auer2002} until time $n$. Then, for all $n \geqslant K$,
\begin{align*}
M_n := \min_{a\in\mathcal{A}}N_a(n) \geqslant f(n),
\end{align*}
where $\left( f(n) : n = K, K+1, ... \right)$ is some deterministic monotone non-decreasing integer-valued sequence satisfying $f(K)=1$ and $f(n)\to\infty$ as $n\to\infty$.
\end{lemma}

\subsection{Proof of Lemma~\ref{lemma:alg4_1}}

Suppose that there exists a sample-path on which some non-empty subset of arms $\mathfrak{A} \subset \mathcal{A}$ receives a bounded number of pulls asymptotically in the horizon of play. Also suppose that $\mathfrak{A}$ is the maximal such subset, i.e., each arm in $\mathcal{A}\backslash\mathfrak{A}$ is played infinitely often asymptotically on said sample-path. This implies that the UCB score of any arm in $\mathcal{A}\backslash\mathfrak{A}$ is at most $1+o\left( \sqrt{\log t} \right)$ at time $t$ (since the empirical mean term remains bounded in $[0,1]$). At the same time, the boundedness hypothesis implies that the UCB score of any arm in $\mathfrak{A}$ is at least $\Omega\left( \sqrt{\log t} \right)$. Thus, for $t$ large enough, UCB scores of arms in $\mathfrak{A}$ will start to dominate those in $\mathcal{A}\backslash\mathfrak{A}$ and the algorithm will end up playing an arm from $\mathfrak{A}$ at some point, thus increasing the cumulative sample-count of arms in $\mathfrak{A}$ by $1$. As $t$ grows further, one can replicate the preceding argument an arbitrary number of times to conclude that $\mathfrak{A}$ receives an unbounded number of pulls on the sample-path under consideration, thereby contradicting the boundedness hypothesis. Therefore, it must be the case that each arm in $\mathcal{A}$ is played infinitely often on \emph{every} sample-path. Consequently, $M_n = \min_{a\in\mathcal{A}}N_a(n) \to \infty$ as $n\to\infty$ on every sample-path. 

Now since $\left( M_n : n = K, K+1, ... \right)$ is an integer-valued process (starting from $M_K=1$) with unit increments (wherever they exist), it follows that on every sample-path, $\tau$, in fact, weakly dominates the stopping time $\tau^\prime$ given by
\begin{align}
\tau^\prime := \inf \left\lbrace m \in \mathbb{N} : \exists\ a,b\in\mathcal{A}, a<b\ \text{s.t.}\ \left\lvert \mathcal{Z}_{a,b} + \sum_{j=1}^{m} \left( X_{a,j} - X_{b,j} \right) \right\rvert < 4\sqrt{m\log m} \right\rbrace.
\end{align}
Therefore, $\mathbb{P}\left( \tau = \infty \right) \geqslant \mathbb{P}\left( \tau^\prime = \infty \right) > \beta_{\delta,K}$; the last inequality follows from Proposition~\ref{propK}. \hfill $\square$

\subsection{Proof of Lemma~\ref{lemma:alg4_2}}

Suppose that $\mathcal{S}_n = \left\lbrace \left( N_a(n) : a\in\mathcal{A} \right)\right\rbrace$ denotes the set of possible sample-count realizations under \texttt{UCB1} when the horizon of play is $n$. Define $f(n) := \min_{\left( N_a(n) : a\in\mathcal{A} \right)\in\mathcal{S}_n}\min_{a\in\mathcal{A}}N_a(n)$. Since $\mathcal{S}_n$ is finite, aforementioned minimum is attained at some $\left( N_a^\ast(n) : a\in\mathcal{A} \right)\in\mathcal{S}_n$. Note that $\left( N_a^\ast(n) : a\in\mathcal{A} \right)$ is not a random vector as it corresponds to a specific set of sample-paths (possibly non-unique) on which $\min_{a\in\mathcal{A}}N_a(n)$ is minimized. Therefore, $f(n) = \min_{a\in\mathcal{A}}N^\ast_a(n)$ is deterministic. We have already established in the proof of Lemma~\ref{lemma:alg4_1} that for each $a\in\mathcal{A}$, $N_a(n) \to \infty$ as $n\to\infty$ on every sample-path. In particular, this also implies $N_a^\ast(n) \to \infty$ as $n\to\infty$. Thus, we have established the existence of a sequence $f(n)$ satisfying the assertions of the lemma. \hfill $\square$

\section{Proof of Theorem~\ref{thm:alg4}}
\label{proof:alg4}

Let $\mathcal{A} := \{1,...,K\}$ be the collection of $K$ arms queried during the first epoch. Consider an arbitrary $l\in\mathbb{N}$ s.t. $l \geqslant K$ and define the following:
\begin{align}
&M_l := \min_{a\in\mathcal{A}} N_a(l), \notag \\
&\tau := \inf \left\lbrace l \geqslant K : \exists\ a,b\in\mathcal{A}, a<b\ \text{s.t.}\ \left\lvert \mathcal{Z}_{a,b} + \sum_{j=1}^{M_l} \left( X_{a,j} - X_{b,j} \right) \right\rvert < 4\sqrt{M_l\log M_l} \right\rbrace, \notag
\end{align}
where $N_a(l)$ denotes the sample count from arm~$a$ under \texttt{UCB1} until time $l$. Note that $\tau$ marks the termination of epoch~$1$. 

Let $R_n$ denote the cumulative pseudo-regret of \texttt{ALG4} after $n$ pulls (superscript $\pi$ suppressed for notational convenience). Let $S_n$ denote the cumulative pseudo-regret of \texttt{UCB1} after $n$ pulls in a $K$-armed bandit with means $\mu_1 > \mu_2 > ... > \mu_K$. Let \texttt{D} denote the event that the $K$ arms queried in epoch~$1$ have distinct types (no two belong to the same type). Let $\tilde{R}_n,\bar{R}_n$ be independently drawn from the same distribution as $R_n$. Then, the evolution of $R_n$ satisfies

\begin{align*}
R_n &\leqslant \mathbbm{1}\{\texttt{D}\}\left[ S_{\min\left(\tau, n\right)} + \tilde{R}_{\left( n-\tau \right)^+} \right] + \mathbbm{1}\{\texttt{D\textsuperscript{c}}\} \left[ \bar{\Delta}\min\left(\tau, n\right) + \bar{R}_{\left( n-\tau \right)^+} \right] \\
&\underset{\mathrm{(\dag)}}{\leqslant} \mathbbm{1}\{\texttt{D}\}\left[ S_{n} + \mathbbm{1}\left\lbrace \tau < n \right\rbrace\tilde{R}_{n} \right] + \mathbbm{1}\{\texttt{D\textsuperscript{c}}\} \left[ \bar{\Delta}\min\left(\tau, n\right) + \bar{R}_{n} \right] \\
&\leqslant \mathbbm{1}\{\texttt{D}\}\left[ S_{n} + \mathbbm{1}\left\lbrace \tau < \infty \right\rbrace\tilde{R}_{n} \right] + \mathbbm{1}\{\texttt{D\textsuperscript{c}}\} \left[ \bar{\Delta}\min\left(\tau, n\right) + \bar{R}_{n} \right],
\end{align*}
where $(\dag)$ follows since \texttt{ALG4} is agnostic to $n$, and hence the pseudo-regret $R_n$ is weakly increasing in $n$. Taking expectations on both sides, one recovers using the independence of $\tilde{R}_n,\bar{R}_n$ that

\begin{align*}
\mathbb{E}R_n &\leqslant \frac{1}{\mathbb{P}\left( \left. \tau = \infty \right\rvert \texttt{D} \right)}\left[ \mathbb{E}S_n + \left( \frac{\bar{\Delta} \mathbb{E}\left[ \left.\min\left(\tau, n\right) \right\rvert \texttt{D\textsuperscript{c}} \right]}{\mathbb{P}(\texttt{D})} \right) \right] \leqslant \frac{1}{\beta_{\delta,K}}\left[ \mathbb{E}S_n + \left( \frac{\bar{\Delta} \mathbb{E}\left[ \left.\min\left(\tau, n\right) \right\rvert \texttt{D\textsuperscript{c}} \right]}{\mathbb{P}(\texttt{D})} \right) \right],
\end{align*}
where $\mathbb{P}(\texttt{D}) = K!\prod_{i=1}^{K}\alpha_i$, and the last inequality follows using Lemma~\ref{lemma:alg4_1} with $\beta_{\delta,K}$ as defined in \eqref{eqn:beta_K}. We know that $\mathbb{E}S_n \leqslant CK\left( \log n/\underline{\Delta} + \bar{\Delta} \right)$ for some absolute constant $C$ \citep{auer2002}. The rest of the proof is geared towards showing that $\mathbb{E}\left[ \left.\min\left(\tau, n\right) \right\rvert \texttt{D\textsuperscript{c}} \right] =  o(n)$.

\subsection{Proof of $\boldsymbol{\mathbb{E}\left[ \left.\min\left(\tau, n\right) \right\rvert \mathtt{D\textsuperscript{c}} \right] =  o(n)}$}

Let $\mathbb{P}_{\mathtt{D\textsuperscript{c}}}(\cdot) := \mathbb{P}(\cdot | \mathtt{D\textsuperscript{c}})$ be the conditional measure w.r.t. the event $\mathtt{D\textsuperscript{c}}$. On $\mathtt{D\textsuperscript{c}}$, there exist two arms in the consideration set $\mathcal{A}$ that belong to the same type. Without loss of generality, suppose that these are indexed by $1,2$. Then,

\begin{align*}
\mathbb{E}\left[ \left.\min\left(\tau, n\right) \right\rvert \mathtt{D\textsuperscript{c}} \right] &\leqslant K + \sum_{k=K+1}^{n}\mathbb{P}_{\mathtt{D\textsuperscript{c}}}\left( \tau \geqslant k \right) \\
&\leqslant K + \sum_{k=K}^{n}\mathbb{P}_{\mathtt{D\textsuperscript{c}}}\left( \tau \geqslant k+1 \right) \\
&= K + \sum_{k=K}^{n}\mathbb{P}_{\mathtt{D\textsuperscript{c}}}\left( \bigcap_{l=1}^{k} \bigcap_{a,b\in\mathcal{A}, a<b}\left\lbrace  \left\lvert \mathcal{Z}_{a,b} + \sum_{j=1}^{M_l} \left( X_{a,j} - X_{b,j} \right) \right\rvert \geqslant 4\sqrt{M_l\log M_l}  \right\rbrace \right) \\
&\leqslant K + \sum_{k=K}^{n}\mathbb{P}_{\mathtt{D\textsuperscript{c}}}\left( \left\lvert \mathcal{Z}_{1,2} + \sum_{j=1}^{M_k} \left( X_{1,j} - X_{2,j} \right) \right\rvert \geqslant 4\sqrt{M_k\log M_k} \right) \\
&= K + \sum_{k=K}^{n}\sum_{m = 1}^{k}\mathbb{P}_{\mathtt{D\textsuperscript{c}}}\left( \left\lvert \mathcal{Z}_{1,2} + \sum_{j=1}^{M_k} \left( X_{1,j} - X_{2,j} \right) \right\rvert \geqslant 4\sqrt{M_k\log M_k}, M_k = m \right) \\
&\underset{\mathrm{(\dag)}}{=} K + \sum_{k=K}^{n}\sum_{m = f(k)}^{k}\mathbb{P}_{\mathtt{D\textsuperscript{c}}}\left( \left\lvert \mathcal{Z}_{1,2} + \sum_{j=1}^{M_k} \left( X_{1,j} - X_{2,j} \right) \right\rvert \geqslant 4\sqrt{M_k\log M_k}, M_k = m \right) \\
&\leqslant K + \sum_{k=K}^{n}\sum_{m=f(k)}^{k}\mathbb{P}_{\mathtt{D\textsuperscript{c}}}\left( \left\lvert \mathcal{Z}_{1,2} + \sum_{j=1}^{m} \left( X_{1,j} - X_{2,j} \right) \right\rvert \geqslant 4\sqrt{m\log m} \right) \\
&\underset{\mathrm{(\ddag)}}{\leqslant} K + 2\sum_{k=K}^{n}\sum_{m=f(k)}^{k}\frac{1}{m^4} \\
&= K + 2\sum_{k=K}^{n}\left( \frac{1}{\left( f(k) \right)^4} + \sum_{m=f(k)+1}^{k}\frac{1}{m^4} \right) \\
&\leqslant K + 2\sum_{k=K}^{n}\left( \frac{1}{\left( f(k) \right)^4} + \frac{1}{3\left( f(k) \right)^3} \right),
\end{align*}
where $(\dag)$ follows from Lemma~\ref{lemma:alg4_2}, and $(\ddag)$ using the Chernoff-Hoeffding bound \citep{hoeffding}. Since $f(k)$ is monotone non-decreasing and coercive in $k$, it follows that $\mathbb{E}\left[ \left.\min\left(\tau, n\right) \right\rvert \mathtt{D\textsuperscript{c}} \right] = o(n)$, where the little-Oh only hides dependence on $K$. \hfill $\square$

\end{document}